\documentclass[%
 aip,
amsmath,amssymb,
preprint,
]{revtex4-2}

\usepackage{comment}
\usepackage{hyperref}
\usepackage{cleveref}

\usepackage{booktabs}

\usepackage{graphicx}
\usepackage{dcolumn}
\usepackage{bm}

\usepackage[utf8]{inputenc}
\usepackage[T1]{fontenc}
\usepackage{etoolbox}

\graphicspath{{./eps/}{./epstopdf_converted/}}

\newcommand*\D {\mathcal{D}}

\newcommand* \disp {\phi}

\newcommand*\z {\varepsilon}

\newcommand* \RevisionDelete[1] {}

\makeatletter
\def\@email#1#2{%
 \endgroup
 \patchcmd{\titleblock@produce}
  {\frontmatter@RRAPformat}
  {\frontmatter@RRAPformat{\produce@RRAP{*#1\href{mailto:#2}{#2}}}\frontmatter@RRAPformat}
  {}{}
}%
\makeatother

\begin{document}



	\title{
        An Inverse Scattering Inspired Fourier Neural Operator for Time-Dependent PDE Learning 
    }
	\author{Rixin Yu}
	\email{rixin.yu@energy.lth.se}
	\affiliation{ 
	Department of Energy Sciences, Lund University, 22100 Lund, Sweden
	}%

	\date{\today}
	
	\begin{abstract}
Learning accurate and stable time-advancement operators for nonlinear partial differential equations (PDEs) remains challenging, particularly for chaotic, stiff, and long-horizon dynamical systems. 
While neural operator methods such as the Fourier Neural Operator (FNO) and Koopman-inspired extensions achieve good short-term accuracy, their long-term stability is often limited by unconstrained latent representations and cumulative rollout errors.
In this work, we introduce an \emph{inverse scattering inspired Fourier Neural Operator} (IS-FNO), motivated by the reversibility and spectral evolution structure underlying the classical inverse scattering transform.
The proposed architecture enforces a near-reversible pairing between lifting and projection maps through an explicitly invertible neural transformation, and models latent temporal evolution using exponential Fourier layers that naturally encode linear and nonlinear spectral dynamics.
We systematically evaluate IS-FNO against baseline FNO and Koopman-based models on a range of benchmark PDEs, including the Michelson-Sivashinsky and Kuramoto-Sivashinsky equations (in one and two dimensions), as well as the integrable Korteweg-de Vries and Kadomtsev-Petviashvili equations. 
The results demonstrate that IS-FNO achieves lower short-term errors and substantially improved long-horizon stability in non-stiff regimes.
For integrable systems, reduced IS-FNO variants that embed analytical scattering structure retain competitive long-term accuracy despite limited model capacity.
Overall, this work shows that incorporating physical structure---particularly reversibility and spectral evolution---into neural operator design significantly enhances robustness and long-term predictive fidelity for nonlinear PDE dynamics.
	\end{abstract}

	\maketitle



\section{Introduction \label{sec:intro}} 

Partial differential equations (PDEs) provide a fundamental mathematical description of spatio-temporal dynamics arising in a wide range of physical systems, including fluid flows, combustion, climate dynamics, and nonlinear wave propagation. Despite their central role, the numerical solution of nonlinear PDEs---especially over long time horizons or in regimes characterized by stiffness, instability, or chaos---remains computationally demanding. High-fidelity solvers often incur prohibitive costs, motivating the development of reduced-order or data-driven alternatives that can deliver accurate and efficient predictions.

Recent advances in machine learning (ML) and artificial intelligence (AI) have substantially reshaped the computational treatment of PDEs. In particular, \emph{operator learning} has emerged as a powerful framework for approximating mappings between infinite-dimensional function spaces. Early data-driven approaches relied on convolutional neural networks (CNNs) \cite{CNN1,CNN2,CNN3,CNN4,CNN5,UNet,ConvPDE}, which interpret discretized PDE solutions as images and learn grid-dependent input-output mappings. While effective in many settings, such approaches are inherently tied to fixed discretizations.

Neural operator methods \cite{GraphKenerlNetwork, kovachki2021neural} overcome this limitation by learning operators directly in function space, enabling discretization-invariant generalization. Representative examples include the Deep Operator Network (DeepONet) \cite{DeepONet} and the Fourier Neural Operator (FNO) \cite{FNO2020}. DeepONet employs a branch-trunk architecture to approximate nonlinear operators, whereas FNO exploits spectral convolution to efficiently represent global interactions in space. Subsequent extensions have incorporated wavelet representations \cite{gupta2021multiwavelet, tripura2023wavelet}, complex geometries \cite{chen2023laplace}, and parameter-dependent operators \cite{YuH2024}.

Among the operators of interest in PDE modeling, the \emph{solution time-advancement operator} plays a particularly important role. Given a solution field $\phi(\boldsymbol{x},t)$ at time $t$, this operator maps it to the solution at a later time $t+\Delta_t$ under prescribed boundary conditions, where $\boldsymbol{x}$ denotes the spatial coordinates and $t$ denotes time. Iterative application of such an operator enables long-horizon prediction of dynamical evolution, which is essential in applications such as unsteady fluid dynamics, flame front propagation, and pattern-forming systems.

Learning accurate and stable time-advancement operators is, however, challenging. While many neural operator models achieve impressive single-step accuracy, small errors can accumulate rapidly during rollout, leading to degraded long-term predictions---particularly for chaotic or unstable PDEs. Prior work has sought to mitigate this issue through recurrent or multi-step training strategies \cite{FNO2020,Yu2023}.
 Nevertheless, long-term robustness remains limited when the latent representations lack physical structure or consistency constraints.

Motivated by \emph{Koopman operator theory} \cite{koopman1931hamiltonian,mezic2013analysis,brunton2022data}, which asserts that nonlinear dynamical systems can be represented as linear evolution in an appropriately lifted observable space, our recent work\cite{Yu_Koop2025} introduced the \emph{Koopman Fourier Neural Operator} (kFNO). By learning a latent linear (or weakly nonlinear) evolution operator, kFNO demonstrated improved performance over standard FNO in modeling unstable flame front dynamics, capturing both short-term evolution and long-term statistical behavior.

In kFNO, the physical-state field is first lifted into a high-dimensional latent function space, evolved forward in time via repeated application of Fourier layers, and then projected back to the physical space. The lifting and projection maps are learned independently and are not constrained to be inverses of one another. From a physical perspective, however, a desirable property of a consistent time-advancement operator is \emph{near reversibility}: in the limit of vanishing time step, the combined lift-evolve-project operation should recover the original state.

A closely related structural principle appears in the classical \emph{inverse scattering transform} (IST) \cite{IST_gardner1967method,IST_ablowitz1981solitons,IST_novikov1984theory}, which provides exact solution formulas for certain integrable nonlinear PDEs. In the IST framework, the solution $\phi(\boldsymbol{x},t_0)$ is mapped to scattering data $\mathcal{S}(t_0)$ via a forward transform,
\begin{equation}
\mathbb{S}:\;\phi(\boldsymbol{x},t_0)\mapsto \mathcal{S}(t_0),
\label{eq:IST_forward}
\end{equation}
the scattering data evolve trivially in time,
\begin{equation}
\mathbb{A}:\;\mathcal{S}(t_0)\mapsto \mathcal{S}(t),
\label{eq:IST_middle}
\end{equation}
and the solution at time $t$ is recovered through the inverse transform,
\begin{equation}
\mathbb{S}^{-1}:\;\mathcal{S}(t)\mapsto \phi(\boldsymbol{x},t).
\label{eq:IST_inverse}
\end{equation}
A key feature of this construction is exact reversibility, $\mathbb{S}^{-1}\circ\mathbb{S}(\phi)=\phi$. A concise overview of IST on an example equation is provided in Appendix~\ref{app:ist}.

Inspired by this structure, we propose an \emph{Inverse Scattering-inspired Fourier Neural Operator} (IS-FNO). The IS-FNO architecture enforces a near-invertible pairing between the lifting and projection networks and introduces an \emph{exponential Fourier layer} to model latent temporal evolution in a manner analogous to the simple spectral dynamics of scattering data. This design embeds physical structure---reversibility and spectral evolution---directly into the neural operator.

We assess the proposed IS-FNO on a suite of benchmark PDEs chosen to span a range of dynamical behaviors and modeling challenges. The Michelson-Sivashinsky and Kuramoto-Sivashinsky equations serve as prototypical examples of unstable and chaotic dissipative systems relevant to flame front and pattern-forming dynamics. The Korteweg-de Vries (KdV) equation provides a canonical integrable model with soliton solutions, while its higher-dimensional extension probes the generalization of operator learning frameworks beyond one-dimensional settings. Together, these benchmarks enable a systematic evaluation of accuracy, stability, and long-horizon robustness.

The remainder of the paper is organized as follows. Section~\ref{sec:OLmethods} formulates the learning problem for solution time-advancement operators. Section~\ref{sec:networks} introduces the proposed architectures. Section~\ref{sec:datasets} describes the benchmark PDE datasets. Section~\ref{sec:results} presents comparative numerical results. Finally, Section~\ref{sec:conclusion} summarizes the main findings and discusses future directions.

\section{Problem Setup for Operator Learning  \label{sec:OLmethods} }

In this section, we outline the problem setup for learning PDE operators, together with a description of recurrent training approaches.

Consider a system governed by a PDE, typically involving mappings between functional spaces. A general operator can be written as
\begin{equation}
    G : \mathcal{V} \to \mathcal{V}'; \quad v(x) \mapsto v'(x'),
\end{equation}
where the input function $v(x)$, with $x \in \mathcal{D}$, belongs to the functional space $\mathcal{V}(\mathcal{D}, \mathbb{R}^{d_v})$. The output function $v'(x')$, with $x' \in \mathcal{D}'$, belongs to $\mathcal{V}'(\mathcal{D}', \mathbb{R}^{d_v'})$, where $\mathcal{D} \subset \mathbb{R}^d$ and $\mathcal{D}' \subset \mathbb{R}^{d'}$ denote the input and output domains, respectively.

In this work, we are primarily interested in the time-advancement operator of the PDE solution, defined as
\begin{equation}
    G: \disp(x, t) \mapsto \disp(x, t_1),
    \label{eq:G_operator}
\end{equation}
where $\disp(x, t)$ denotes the PDE solution at time $t$, and $t_j = t + j\Delta_t$ for $j=1,2,\dots$ represents future time instances separated by a time step $\Delta_t$. For simplicity, we assume that the input and output share the same domain and codomain, i.e.\ $\mathcal{D}' = \mathcal{D}$, $\mathcal{V}' = \mathcal{V}$, $d' = d$, and $d_v' = d_v$, with periodic boundary conditions on $\mathcal{D}$.

In some cases, it is preferable to work with a related operator that outputs solutions over multiple consecutive steps:
\begin{equation}
    \bar{G}: \mathcal{V} \to \mathcal{V}^n; \quad 
    \disp(x, t) \mapsto (\disp(x, t_1), \disp(x, t_2), \dots, \disp(x, t_n)),
    \label{eq:Gn_operator}
\end{equation}
where $\mathcal{V}^n$ denotes the Cartesian product of $n$ copies of $\mathcal{V}$. These operators satisfy the relationship
\begin{equation}
    \bar{G} = (G, G^2, \dots, G^n),
    \label{eq:Gn_G}
\end{equation}
where the superscript denotes iterative composition: $G^n = G \circ \cdots \circ G$ ($n$ times).

To approximate the operators $G$ and $\bar{G}$ using neural networks, we introduce the parameterized models
\begin{equation}
    \mathcal{G}: \mathcal{V} \times \Theta \to \mathcal{V}, 
    \text{   or     }
    \mathcal{G}_{\theta}: \mathcal{V} \to \mathcal{V},
\end{equation}
and
\begin{equation}
    \bar{\mathcal{G}}: \mathcal{V} \times \Theta \to \mathcal{V}^n, 
    \text{   or     }
    \bar{\mathcal{G}}_{\theta}: \mathcal{V} \to \mathcal{V}^n,
\end{equation}
where $\theta \in \Theta$, denoting the space of trainable parameters. The learning objective is to find $\hat{\theta} \in \Theta$ such that $\mathcal{G}_{\hat{\theta}} \approx G$ or $\bar{\mathcal{G}}_{\hat{\theta}} \approx \bar{G}$.

Our previous studies \cite{Yu2023, YuH2024, YHN2024} focused on learning the single-step advancement operator $G$. Starting from an initial solution $\disp(x, t)$, recurrent application of the learned operator $\mathcal{G}_{\hat{\theta}}$ enables long-range rollout predictions by iteratively feeding each output back into the model. A well-learned operator should yield accurate short-term predictions. Although long-term predictions generally diverge for chaotic PDEs, it is still desirable for them to reproduce key statistical properties of the underlying nonlinear system.

To promote numerical stability, the works \cite{Yu2023, YuH2024, YHN2024} employed a one-to-many training strategy, in which $\mathcal{G}_{\theta}$ was trained to predict multiple successive steps from a single input. Specifically, given a dataset
\[
    \{\, v_j,\; (G^1 v_j, \dots, G^n v_j) \,\}_{j=1}^Z
\]
consisting of $Z$ input-output pairs arranged in a $1$-to-$n$ format, the loss function is
\begin{equation}
    \hat{J}
    = 
    \mathbb{E}_{v \sim \chi}
    \left[
        C\left((\mathcal{G}_{\theta}^1 v, \dots, \mathcal{G}_{\theta}^n v),\; (G^1 v, \dots, G^n v)\right)
    \right],
    \label{eq:loss}
\end{equation}
where $v$ is sampled from a distribution $\chi$ over the input space, and $C$ denotes  the relative mean squared error
\[
    C(a,b) = \frac{\|a - b\|_2}{\|b\|_2}.
\]
Minimizing this loss,
\[
    \hat{\theta} = \arg\min_{\theta \in \Theta} \hat{J},
\]
yields the trained operator $\mathcal{G}_{\hat{\theta}}$.

In our earlier work on Koopman-inspired neural operators \cite{Yu_Koop2025}, we proposed an alternative approach that directly learns the multi-step operator $\bar{G}$ through the network $\bar{\mathcal{G}}_\theta$. In the present work, we continue to focus on learning $\bar{G}$. Using the same training dataset, the loss function becomes
\begin{equation}
    \hat{J}
    =
    \mathbb{E}_{v \sim \chi}
    \left[
        C\left(\bar{\mathcal{G}}_{\theta} v,\; (G^1 v, \dots, G^n v)\right)
    \right],
    \label{eq:loss_extended}
\end{equation}
and minimizing it produces the trained model $\bar{\mathcal{G}}_{\hat{\theta}}$.

After training, recurrent application of $\bar{\mathcal{G}}_{\hat{\theta}}$ enables long-horizon predictions by repeatedly feeding the last predicted state (i.e.\ the $n$th component) back as the next input. More precisely, given an initial input $\disp(x,t)$, the predicted state at time $t + j\Delta_t$ for integer $j \ge 1$ is obtained as
\begin{equation}
    \bigl( \bar{\mathcal{G}}_{\hat{\theta}}^{\lceil j/n \rceil}\, \disp(x,t) \bigr)\,[\, j \bmod n \,].
\end{equation}

\section{Operator Learning Methods}
\label{sec:networks}

This section describes the operator-learning architectures used in this work. We begin with a brief review of the baseline Fourier Neural Operator (FNO) \cite{FNO2020}, followed by the Koopman-theory-inspired extension (kFNO) \cite{Yu_Koop2025}. We then introduce the proposed inverse scattering inspired Fourier Neural Operator (IS-FNO).

\subsection{Baseline Fourier Neural Operator (FNO) }

FNO \cite{FNO2020} learns infinite-dimensional operators by parameterizing integral kernels in Fourier space. When applied to learning the single-step time-advancement operator $G$, FNO models the mapping $\phi(x,t) \mapsto \phi(x,t_1)$ through the composition
\begin{equation}
    \mathcal{G}_\theta \big|_{\text{FNO}}
    = P \circ H' \circ L ,
    \nonumber
\end{equation}
where $L$ is a lifting map, $H'$ is a hidden evolution map, and $P$ is a projection map.

The lifting map $L : \mathcal{V} \to \mathcal{V}_\z$ embeds the input field into a higher-dimensional feature space, $\phi(x,t) \mapsto \z_0(x)$, where $\mathcal{V}_\z := \mathcal{V}_\z(\mathbb{R}^d; \mathbb{R}^{d_\z})$ with $d_\z > d_v$. A common implementation is a single linear layer,
\begin{equation}
    \z_0(x) = \boldsymbol{w}_L \, \phi(x,t) + \boldsymbol{b}_L ,
    \label{eq:L}
\end{equation}
where $\boldsymbol{w}_L \in \mathbb{R}^{d_v \times d_\z}$ and $\boldsymbol{b}_L \in \mathbb{R}^{d_\z}$ are trainable parameters.

The hidden evolution map $H' : \mathcal{V}_\z \to \mathcal{V}_\z$ performs a sequence of updates $\z_0 \mapsto \z_1 \mapsto \cdots \mapsto \z_L$. Each update $\z_l \mapsto \z_{l+1}$ is implemented via a Fourier layer of the form
\begin{equation}
    \z_{l+1}
    = \alpha \z_l
    + \sigma\!\left(
        \boldsymbol{w}_{\mathcal{F}} \, \z_l
        + \boldsymbol{b}_{\mathcal{F}}
        + \mathcal{F}^{-1}\!\left\{
            \boldsymbol{r}^* \cdot \mathcal{F}\{\z_l\}
        \right\}
    \right),
    \label{eq:FourierLayer}
\end{equation}
where $\boldsymbol{w}_{\mathcal{F}} \in \mathbb{R}^{d_\z \times d_\z}$ and $\boldsymbol{b}_{\mathcal{F}} \in \mathbb{R}^{d_\z}$ are learnable parameters, $\sigma$ is a pointwise nonlinearity, and $\mathcal{F}$ and $\mathcal{F}^{-1}$ denote the Fourier transform and its inverse. The coefficient $\alpha \in \{0,1\}$ controls the residual connection: $\alpha=0$ disables it, while $\alpha=1$ enables a full skip connection.

The tensor $\boldsymbol{r}^* \in 
\mathbb{C}^{(2\kappa_1^{\max}-1)\times \cdots \times \kappa_d^{\max}\times d_\z \times d_\z}$
contains complex-valued learnable weights that mix Fourier modes across the feature dimension accroding to
\begin{equation}
\bigl(\boldsymbol{r}^* \cdot \mathcal{F}\{\z\}\bigr)_{\boldsymbol{\kappa},\, i}
=
\sum_{j=1}^{d_\z}
(\boldsymbol{r}^*)_{\boldsymbol{\kappa},\, i,j}\,
\mathcal{F}\{\z\}_{\boldsymbol{\kappa},\, j},
\qquad i=1,\ldots,d_\z,
\label{eq:mix_of_fourier_mode}
\end{equation}
where $\boldsymbol{\kappa}=(\kappa_1,\ldots,\kappa_d)$ denotes the multi-dimensional wave-number index.
For real-valued $\z(x)$, conjugate symmetry is enforced by restricting independent modes to
$\kappa_m\in(-\kappa_m^{\max},\kappa_m^{\max})$ for $m<d$ and $\kappa_d\in[0,\kappa_d^{\max})$.
Fourier modes outside a user-prescribed cutoff $\boldsymbol{\kappa}^{\max}$ are truncated to zero.

Finally, the projection map $P : \mathcal{V}_\z \to \mathcal{V}$ maps the final hidden state back to the physical space, $\z_L(x) \mapsto \phi(x,t_1)$. In practice, $P$ is typically implemented as a two-layer perceptron.

\subsection{Koopman Fourier Neural Operator (kFNO)}

The kFNO architecture \cite{Yu_Koop2025} generalizes the baseline FNO by learning the $n$-step solution-advancement operator $\bar{G}$. This is achieved through the composite mapping
\begin{equation}
    \bar{\mathcal{G}}_\theta \big|_{\text{kFNO}}
    = \bar{P} \circ \bar{Q} \circ \bar{K} \circ H \circ L ,
    \nonumber
\end{equation}
as illustrated in Fig.~\ref{fig:invFNO}.

Following the baseline FNO, the input field undergoes a two-stage lifting,
\[
    \phi(x,t) \;\mapsto\; \z_0(x) \;\mapsto\; \z'_0(x),
\]
where $\z'_0 \in \mathcal{V}_\z$ is obtained via the lifting map $L$ followed by a hidden evolution map $H : \mathcal{V}_\z \to \mathcal{V}_\z$, analogous to $H'$.

The Koopman-inspired extension introduces the operator
\[
    \bar{K} : \mathcal{V}_\z \to \mathcal{V}_\z^{\,n};
    \qquad
    \z'_0(x) \mapsto (\z'_1(x), \ldots, \z'_n(x)),
\]
where future hidden states are generated via repeated application of a single advancement operator $A : \mathcal{V}_\z \to \mathcal{V}_\z$:
\[
    \z'_j(x) = A^{\,j} \z'_0(x), \qquad j = 1,\ldots,n.
\]
This recursive structure enables kFNO to approximate multi-step temporal evolution directly in the lifted feature space. Prevous work \cite{Yu_Koop2025} recommends $A$ to be implemented as a stack of Fourier layers.

The map 
\[
    \bar{Q}: \mathcal{V}_\z^{\,n} \to \mathcal{V}_\z^{\,n};
    \qquad
    (\z'_1,\ldots,\z'_n) \mapsto (\z^\dagger_1,\ldots,\z^\dagger_n),
\]
applies additional transformations to each intermediate hidden state. A simple and effective choice \cite{Yu_Koop2025} is to apply a shared operator $Q : \mathcal{V}_\z \to \mathcal{V}_\z$ independently to each  $\z'_j$.

Finally, the projection map 
\[
    \bar{P}: \mathcal{V}_\z^{\,n} \to \mathcal{V}^n
\]
maps each refined hidden state $\z^\dagger_j$ to the corresponding physical output $\phi(x,t_j)$, typically by independently applying the same projection operator $P$.

\subsection{Inverse Scattering inspired Fourier Neural Operator (IS-FNO)}
\label{sec:IS-FNO} 

\begin{figure*}
	\centerline{
		\includegraphics[width=1\linewidth]{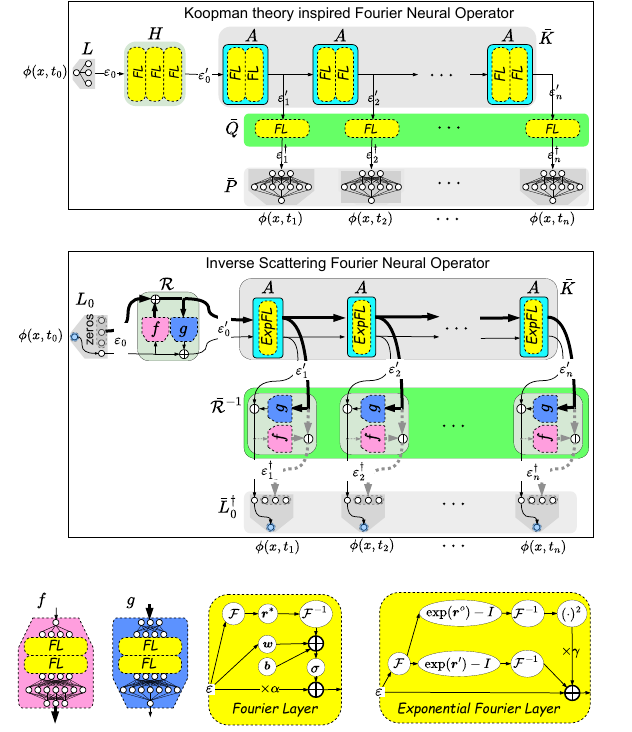}
	}
	\caption{
		\label{fig:invFNO}
    Schematic comparison of inverse scattering based and Koopman inspired Fourier Neural Operator architectures.
	}
\end{figure*}

We now introduce the proposed IS-FNO, which learns the $n$-step solution-advancement operator $\bar{G}$ through the composition (see Fig.~\ref{fig:invFNO})
\begin{equation}
    \bar{\mathcal{G}}_\theta \big|_{\text{IS-FNO}}
    =
    \bar{L}_0^\dagger \circ \bar{\mathcal{R}}^{-1} \circ \bar{K} \circ \mathcal{R} \circ L_0 .
    \nonumber
\end{equation}

The first lifting map $L_0 : \mathcal{V} \to \mathcal{V}_\z ; \phi(x,t) \mapsto \z_0(x)$ is implemented by stacking $(d_\z - d_v)$ zero-valued channels onto the input field. This corresponds to using the linear lifting map \eqref{eq:L} with
\begin{equation}
    \boldsymbol{b}_L = \boldsymbol{0}, \qquad
    (\boldsymbol{w}_L)_{ij} = \delta_{ij},
    \label{eq:L_to_L_star}
\end{equation}
i.e., an identity embedding padded with zeros.

The second map $\mathcal{R} : \mathcal{V}_\z \to \mathcal{V}_\z; \z_0(x) \mapsto \z'_0(x)$, is constructed as a reversible residual network (RevNet) \cite{gomez2017reversible}.  
The input and output are split along the co-dimension direction:
\[
    \z_0 = a \cup b,
    \qquad
    \z_0' = a' \cup b',
\]
with  
\[
    a, a' \in \mathcal{V}_{\z_a}(\mathbb{R}^d;\mathbb{R}^{d_{\z_a}}),
    \qquad
    b, b' \in \mathcal{V}_{\z_b}(\mathbb{R}^d;\mathbb{R}^{d_{\z_b}}),
\]
where $1 \le d_{\z_a} \le d_\z - 1$ and $d_{\z_b} = d_\z - d_{\z_a}$.
The map takes the form
\begin{align}
    a' &= a + g\!\left(b + f(a)\right), \nonumber \\
    b' &= b + f(a),
    \label{eq:R}
\end{align}
with
\[
    f : \mathcal{V}_{\z_a} \to \mathcal{V}_{\z_b},
    \qquad
    g : \mathcal{V}_{\z_b} \to \mathcal{V}_{\z_a}.
\]
Each of the sub-maps $f$ and $g$ is implemented in three parts:  
(i) an “adapter’’ that maps the input to an intermediate functional space $\mathcal{V}_{\z_*}(\mathbb{R}^d;\mathbb{R}^{d_{\z_*}})$ via a single linear layer (or identity if the co-dimension already matches),  
(ii) a few stacked Fourier layers operating in the fixed co-dimension $d_{\z_*}$, and  
(iii) a two-layer perceptron that adjusts the co-dimension, first raising it to a large number (we use 128) and then reducing it to the target dimension ($d_{\z_b}$ for $f$ and $d_{\z_a}$ for $g$).

Importantly, $\mathcal{R}$ admits an exact inverse  
$\mathcal{R}^{-1} : \mathcal{V}_\z \to \mathcal{V}_\z ;  (a' \cup b') \mapsto (a \cup b)$,  
given by
\begin{align}
    a &= a' - g(b'), \nonumber \\
    b &= b' - f\!\left(a' - g(b')\right).
    \label{eq:R_inv}
\end{align}

As in kFNO, the Koopman map
$    \bar{K} : \mathcal{V}_\z \to \mathcal{V}_\z^n;
    \z'_0(x) \mapsto (\z'_1(x), \ldots, \z'_n(x))
$
is realized by repeated application of an advancement operator $A : \mathcal{V}_\z \to \mathcal{V}_\z$, so that $\z'_j(x) = A^j \z'_0(x)$ with $A: \z'_j \mapsto \z'_{j+1}$. 
While $A$ can be implemented using vanilla Fourier layers, we propose a more effective \textit{exponential Fourier layer}:
\begin{equation}
    \z'_{j+1}
    = \z'_j
    + \mathcal{F}^{-1}\!\left\{ [\exp(\boldsymbol{r}') - \boldsymbol{I}] \cdot \mathcal{F}\{\z'_j\} \right\}
    + \gamma \left(
       \mathcal{F}^{-1}\!\left\{ [\exp(\boldsymbol{r}^o) - \boldsymbol{I}] \cdot \mathcal{F}\{\z'_j\} \right\}
    \right)^2,
    \label{eq:ExpFourierLayer}
\end{equation}
where $\boldsymbol{r}', \boldsymbol{r}^o \in \mathbb{C}^{(2\kappa_1^{\max}-1) \times \ldots \times \kappa_d^{\max}  \times d_\z \times d_\z}$ are learnable parameters, the matrix exponential is applied over the last two dimensions, 
$\boldsymbol{I}$ is the identity matrix broadcast to match the appropriate shape, and 
Fourier modes with frequencies above $\boldsymbol{\kappa}^{\max}$ are truncated.

This layer may be interpreted as sequentially updating the scattering data as
$
    \mathcal{F}\{\z'_j\} \mapsto \mathcal{F}\{\z'_{j+1}\} .
$
The parameter $\gamma \in \{0,1\}$ is user-specified.  
When $\gamma = 0$, the layer reduces to a purely linear evolution:
\begin{equation}
    \mathcal{F}\{\z'_j\}
    = \exp(j\boldsymbol{r}') \cdot \mathcal{F}\{\z'_0\},
    \qquad
    j = 0,1,\ldots,n,
    \label{eq:linearExpFourierLayer}
\end{equation}
of which the KdV scattering relation~\eqref{eq:kdv-scatter} is a special case.  
When $\gamma = 1$, the additional quadratic term accommodates nonlinear effects.

The fourth map $\bar{\mathcal{R}}^{-1} : \mathcal{V}_\z^n \to \mathcal{V}_\z^n$ applies $\mathcal{R}^{-1}$ to each time step:  
\[
    \z'_j(x) \mapsto \z^\dagger_j(x),
    \qquad j = 1,\ldots,n.
\]

Finally, $\bar{L}_0^\dagger$ maps the multi-step hidden states back to the solution space, producing  
$(\phi(x,t_1), \ldots, \phi(x,t_n))$.  
Each update $\z^\dagger_j(x) \mapsto \phi(x,t_j)$ is achieved via the simple dimension-truncation map  
$L_0^\dagger : \mathbb{R}^{d_\z} \to \mathbb{R}^{d_v}$,  
which extracts the first $d_v$ components of $\z^\dagger_j(x)$.  
Thus, the remaining $(d_\z - d_v)$ co-dimensions of $\z^\dagger_j(x)$ need not be computed.  
If we further set $d_{\z_a} = d_v$, the second part of $f$ in~\eqref{eq:R_inv} is unnecessary and may be skipped to reduce computation.

We conclude by noting a generalization of the pair $(L_0, L_0^\dagger)$.  
The zero-stacking implementation of $L_0$ can be seen as a special case of the learnable lifting map~\eqref{eq:L} through using  \eqref{eq:L_to_L_star}.
If $L_0$ is instead implemented as a learnable $L$, the paired projection becomes a pseudo-inverse:
\[
    \phi(x,t_j)
    = \boldsymbol{w}_L^\dagger \bigl( \z^\dagger_j(x) - \boldsymbol{b}_L \bigr),
\]
where
\[
    \boldsymbol{w}_L^\dagger = (\boldsymbol{w}_L^T \boldsymbol{w}_L)^{-1} \boldsymbol{w}_L^T.
\]

\subsection{Models and Variants}
\label{sec:modelvariants}

The kFNO and IS-FNO architectures admit several interchangeable design choices, resulting in a family of model variants. Below, we summarize all variants evaluated in this work.

\vspace{0.5em}
\noindent\textbf{Koopman Fourier Neural Operator variants.}
Following \cite{Yu_Koop2025}, the intermediate maps $Q$ and $H$ are implemented using 
one and three-stacked vanilla Fourier layers, respectively. Variants differ primarily in the implementation of the advancement operator $A$:
\begin{itemize}
    \item \texttt{kFNO$^*$}: $A$ is implemented using two stacked vanilla Fourier layers (Eq. \ref{eq:FourierLayer}, $\alpha=1$);
    \item \texttt{kFNO$^o$}: $A$ is implemented using the nonlinear exponential Fourier layer (Eq.~\ref{eq:ExpFourierLayer}, $\gamma=1$);
    \item \texttt{kFNO$'$}: $A$ is implemented using the linear exponential Fourier layer (Eq.~\ref{eq:ExpFourierLayer}, $\gamma=0$).
\end{itemize}

\vspace{0.5em}
\noindent\textbf{Baseline FNO.}
A single-step operator learning model is recovered as a special case of the Koopman architecture:
\begin{itemize}
    \item \texttt{baseline FNO}: obtained from \texttt{kFNO$^*$} by setting the rollout horizon to $n=1$, in which case the composed map $\bar{P}\circ\bar{Q}\circ\bar{K}\circ H\circ L$ reduces to the standard $P\circ H' \circ L$ through absorption of $Q$ and $K$ into $H' = Q\circ K \circ H$.
\end{itemize}
In all Koopman-based models (including baseline FNO), the vanilla Fourier layers in $H$, $A$, and $Q$ use skip connections with coefficient $\alpha=1$.

\vspace{0.5em}
\noindent\textbf{Inverse Scattering FNO variants.}
Within the IS-FNO framework, we study the parallel set of variants:
\begin{itemize}
    \item \texttt{IS-FNO$^*$}: $A$ uses two stacked vanilla Fourier layers ($\alpha=1$ with skip connection).
    \item \texttt{IS-FNO$^o$}: $A$ uses the nonlinear exponential Fourier layer ($\gamma=1$);
    \item \texttt{IS-FNO$'$}: $A$ uses the linear exponential Fourier layer ($\gamma=0$);
\end{itemize}

To more accurately evaluate performance on the 1d KdV equation---whose scattering data evolve exactly as $\propto e^{-\kappa^3 t}$ (Eq.~\ref{eq:kdv-scatter})---we further introduce two KdV-specific IS-FNO variants.  
These employ a \emph{modified} linear exponential Fourier layer in $A$, in which the frequency-dependent weights are parameterized as
\begin{equation}
\boldsymbol{r}'(\kappa) = \boldsymbol{r}'' \left( \frac{\kappa}{\kappa_{\max}} \right)^p ,
\qquad \kappa = 0,\ldots,\kappa_{\max},
\end{equation}
where $\boldsymbol{r}'' \in \mathbb{C}^{d_\z \times d_\z}$ is a reduced-size learnable weight matrix and $p$ is an exponent controlling spectral scaling.
The two variants considered are:
\begin{itemize}
    \item \texttt{IS-FNO$'_{\kappa 3}$}: exponent fixed to $p=3$, matching the analytical KdV scattering law;
    \item \texttt{IS-FNO$'_{\kappa}$}: exponent $p$ treated as a learnable parameter.
\end{itemize}

Finally, across all IS-FNO variants, the Fourier layers in sub-maps $f$ and $g$ omit skip connections ($\alpha=0$) to avoid redundant residual pathways already captured by the reversible pair $( \mathcal{R}, \mathcal{R}^{-1} )$.

\section{Governing Equations and Training Datasets}
\label{sec:datasets}

This section presents the governing PDEs and the construction of training datasets used to evaluate the proposed operator-learning models.  
We focus on three families of nonlinear equations exhibiting distinct dynamical regimes:

\begin{itemize}
    \item the Michelson--Sivashinsky (MS) and Kuramoto--Sivashinsky (KS) equations, which model flame-front evolution driven by Darrieus--Landau (DL) \cite{DARRIEUS1938UNPB,landau1988theory} and diffusive--thermal (DT) \cite{zeldovich1944selected,sivashinsky1977diffusional} instabilities;
    \item the Korteweg--de Vries (KdV) equation\cite{kdv1895} and its two-dimensional extension, the Kadomtsev--Petviashvili (KP) equation\cite{KP_kadomtsev1970stability}, which describe weakly nonlinear dispersive waves and soliton interactions.
\end{itemize}

High-resolution pseudo-spectral solvers are used to generate all datasets, covering eight distinct cases:  
five one-dimensional configurations (MS and KS at two parameter settings each, and the KdV equation) and three two-dimensional configurations (MS, KS, and KP equation).  
These problems collectively span chaotic dynamics, stiff cusp formation, multiscale flame-front evolution, and integrable soliton behavior, providing a comprehensive benchmark for assessing operator-learning architectures.

\subsection{Governing Equations for Flame-Front Instabilities}

The spatiotemporal evolution of a statistically planar flame front is described by its displacement $\hat{\phi}(\hat{x},\hat{t}) : \mathbb{R}^d \times \mathbb{R} \rightarrow \mathbb{R}$.  
The full $d$-dimensional Sivashinsky equation \cite{SivaEq} reads
\begin{equation}
\partial_{\hat{t}}  \hat{\phi}
+ \tfrac{1}{2}\left(\nabla\hat{\phi}\right)^2
=
-4(1+\mathrm{Le}^*)^2 \nabla^4 \hat{\phi}
- \mathrm{Le}^* \nabla^2 \hat{\phi}
+ (1-\Omega)\,\Gamma(\hat{\phi}),
\label{eq:MKS_org}
\end{equation}
where $\Gamma$ is a linear nonlocal operator,
\[
\Gamma(\hat{\phi})
= \mathcal{F}^{-1}\!\left(|\kappa|\,\mathcal{F}(\hat{\phi})\right),
\]
and $\Omega$ is the density ratio between products and reactants.  
Here $\nabla$, $\nabla^2$, and $\nabla^4$ denote the gradient, Laplacian, and biharmonic operators.  $\mathrm{Le}^*$ is the effective Lewis number, representing the ratio of thermal to mass diffusivity in the deficient reactant.

With suitable nondimensionalization\cite{YHN2024} of time $t \sim \hat{t}$, space $x \sim \hat{x}$, and the flame displacement $ \phi \sim \hat{\phi} $, the Sivashinsky equation reduces to two limiting cases:

\begin{enumerate}
    \item \textbf{Michelson--Sivashinsky (MS)}\cite{michelson1977nonlinear} (DL instability):
    \begin{equation}
    \frac{1}{\tau} \partial_t \phi
    + \frac{1}{2\beta^2}(\nabla\phi)^2
    =
    \frac{1}{\beta^2}\nabla^2\phi
    + \frac{1}{\beta}\Gamma(\phi),
    \label{eq:MS}
    \end{equation}

    \item \textbf{Kuramoto--Sivashinsky (KS)}\cite{kuramoto1978diffusion} (DT instability):
    \begin{equation}
    \partial_t \phi
    + \frac{1}{2\beta^2}(\nabla\phi)^2
    =
    -\frac{1}{\beta^2}\nabla^2\phi
    -\frac{1}{\beta^4}\nabla^4\phi.
    \label{eq:KS}
    \end{equation}
\end{enumerate}

We consider periodic domains $\mathcal{D}=[0,2\pi]^d$.  
The parameter $\beta$ corresponds to the largest unstable wave number according to linear stablity analysis;  $\tau=\beta/10$ adjusts the time scales of DL instality \cite{YHN2024}.

\paragraph{Known dynamical behavior.}
The KS equation exhibits spatiotemporal chaos for sufficiently large $\beta$ and is widely used as a benchmark problem in PDE learning.  
The MS equation admits an exact pole-decomposition formulation \cite{Thual_Frisch_Henon_poledecomp}, reducing the PDE to a finite-dimensional ODE system.  
At small $\beta$, MS solutions converge to a stable ``giant-cusp’’ profile.  
At large $\beta$, the cusp tip becomes highly curved and extremely sensitive to round-off noise, leading to persistent small-scale wrinkles that convect along the front.  
For very large $\beta$ (e.g., $\beta\gtrsim50$), resolving peak curvature requires significantly finer meshes \cite{Yu2023}.  
Additional background may be found in \cite{Vaynblat_matalon_polestability1, Vaynblat_matalon_polestability2, Olami_noise, denet2006stationary, Kupervasser_pole_book, Karlin2002cellular, Creta2020propagation, CRETA2011INST, rasool2021effect, YBB15PRE}.

\subsection{Governing Equations for Nonlinear Wave Dynamics}

The 1d Korteweg--de Vries (KdV) equation \cite{kdv1895},
\begin{equation}
\partial_t \phi + 6\phi \partial_x \phi + \partial_x^3 \phi = 0,
\label{eq:kdv}
\end{equation}
describes weakly nonlinear, weakly dispersive waves and is a classical integrable system.  
Its 2d extension, the Kadomtsev--Petviashvili (KP) equation \cite{KP_kadomtsev1970stability} of type II, is
\begin{equation}
\partial_{x_1}\!\left( \partial_t \phi + 6\phi \partial_{x_1}\phi + \partial_{x_1}^3 \phi \right)
+  \partial_{x_2}^2 \phi = 0,
\label{eq:2dkdv}
\end{equation}
for which $x=(x_1,x_2)$.
We consider periodic domains $\mathcal{D}= [0,l_{\mathrm{KDV}}]^d$ with $l_{\mathrm{KDV}}=20$.  
Well-posedness of KP requires intial condition  $\phi_0(x) := \phi(x,t_0)$ to satisfy the constraint
\begin{equation}
\int_0^{l_{\mathrm{KDV}}} \left( \partial_{x_2}^2 \phi_0 \right)  \mathrm{d}x_1 = 0.
\label{eq:2dkdv_init}
\end{equation}

\paragraph{IST structure (summary).}
As outlined in Appendix~\ref{app:ist}, the 1d KdV equation is solvable via the inverse scattering transform (IST) \cite{IST_gardner1967method,IST_ablowitz1981solitons,IST_novikov1984theory}.  
The initial field serves as a potential in a stationary Schrödinger problem \eqref{eq:schr}; discrete eigenvalues correspond to solitons, and continuous spectra to dispersive radiation.  
Time evolution is encoded through the isospectral Lax pair \cite{lax1968integrals}, and the solution is reconstructed via the Gelfand--Levitan--Marchenko equation \cite{GLM_gel1951determination,GLM_marchenko2011sturm}, \eqref{eq:GLM}.

The one-soliton solution is
\begin{equation}
\phi(x,t)
= \frac{s}{2}\,\mathrm{sech}^2\!\left(\frac{\sqrt{s}}{2}(x - st)\right),
\label{eq:1dsol}
\end{equation}
whose associated Schrödinger operator has one discrete eigenvalue $\lambda=-s/2$.

\subsection{Training Dataset Construction}
\label{sec:training_data}

All datasets are generated using a pseudo-spectral solver with integrating factors for linear terms and a fourth-fifth order Runge--Kutta scheme for nonlinear terms \cite{kassam2005fourth}.

\subsubsection{One-Dimensional Datasets}
\label{sec:1ddata}

We construct five 1d datasets: MS and KS at $\beta=10$ and $40$, and the KdV equation.  All 1d simulations use 256 grid points.

\paragraph{MS and KS dynamics.}
For KS at both values of $\beta$, long-time solutions are chaotic.  For MS at $\beta=10$, the solution relaxes to a stable giant-cusp profile, whereas at $\beta=40$ the dynamics become highly sensitive to round-off noise, producing persistent small-scale wrinkles riding on a large cusp.  
Note that at even larger $\beta$, the required resolution exceeds the 256-mode spectral discretization, and 512 modes are needed for numerical stability \cite{Yu2023}.

\paragraph{MS and KS datasets.}
For each MS and KS case we generate:
\begin{itemize}
    \item \textbf{250 short} trajectories on $0\le t \le 75$, with 500 snapshots and $\Delta_t = 0.15$;
    \item \textbf{one long} trajectory on $0\le t \le 18{,}750$, containing 125{,}000 snapshots.
\end{itemize}
Initial conditions are random physical-space perturbations 
\begin{equation}
\phi_0(x_j) \sim \mathcal{U}(0,0.03),
\label{eq:MS_KS_init}
\end{equation}
where $x_j$ are the grid points and $\mathcal{U}(a,b)$ denotes the uniform distribution on $[a,b]$.
For more details about MS and KS datasets, the reader is referred to \cite{Yu_Koop2025}.

\paragraph{KdV dataset.}
We generate 350 sequences on $0\le t\le1.22$, each with 500 snapshots at $\Delta_t=0.0024$.  
Two classes of initial conditions are used:

\begin{enumerate}
    \item \textbf{Low-wavenumber randomization:}  
    250 sequences are initialized by prescribing Fourier coefficients as:
    \begin{equation}
    \mathcal{F}_\kappa\{\phi_0\} = \xi_\kappa e^{i\theta_\kappa},
    \label{eq:init_lowwavenumber}
    \end{equation}
    where $\theta_\kappa \in [0,2\pi)$ and $\xi_\kappa \in [0,256/2]$ for $\kappa \le 9$, and zero otherwise.
    
    \item \textbf{Random soliton superposition:}  
    The remaining 100 sequences are initialized by superimposing $M \in [3,8]$ randomly selected soliton solutions (Eq.~\ref{eq:1dsol}):
    \begin{equation}
    \phi_0(x) = \sum_{j=1}^M \frac{s_j}{2}\,\mathrm{sech}^2\!\left[ \frac{\sqrt{s_j}}{2}(x-c_j) \right],
    \label{eq:init_1dsol}
    \end{equation}
    where $s_j \in [0.4,2]$ and $c_j \in [0,l_{\mathrm{KDV}}]$, with periodic wrapping when $x-c_j$ falls outside $\mathcal{D}$.  
    If the initial wave packets do not overlap, each soliton propagates at its own speed until interactions occur.
\end{enumerate}

\subsubsection{Two-Dimensional Datasets}
\label{sec:2ddata}

We construct three 2d datasets: MS and KS at $\beta=15$, and the KDV/KP equation. All simulations use a $128^2$ grid.

For 2d MS and KS, we generate 45 sequences over $0\le t\le435.12$ with 5880 snapshots at inteval $\Delta_t = 0.074$.
Initial conditions use the low-wavenumber randomization \eqref{eq:init_lowwavenumber} with $\xi_\kappa\in[0,0.01]$ for $|\kappa|<9$.  
The MS and KS datasets were used in previous work 
\cite{Yu_Koop2025,Yu_Koop2025_2} to evaluate performance of several (kFNO-based and CNN-based) model variants, showing that randomization in either Fourier or physical space yields comparable learning performance.

For the KDV/KP dataset, we generate 350 sequences using the same time horizon and timestep as the 1d KdV dataset.  
Initial conditions follow the randomized spectrum \eqref{eq:init_lowwavenumber} with $\xi_\kappa\in[0,128^2/15]$ for $|\kappa|\le9$, and additionally satisfy the KP constraint \eqref{eq:2dkdv_init}.

\paragraph{Validation sets.}  
For each of the eight total datasets (five 1d and three 2d), a validation dataset is constructed, comprising at least 10\% of the corresponding training data.

\section{Results and Discussion}
\label{sec:results}

This section evaluates the performance of the proposed inverse scattering inspired neural operators in comparison with Koopman-based models and the baseline \texttt{FNO}, spanning nine model variants described in Sec.~\ref{sec:modelvariants}. All experiments cover the eight datasets introduced in Sec.~\ref{sec:datasets}, encompassing both one- and two-dimensional PDEs and including integrable, chaotic, and noise-sensitive stiff regimes.

Among the nine variants in Sec.~\ref{sec:modelvariants}, seven are benchmarked across all five one-dimensional datasets (1d MS and KS at two values of $\beta$, and 1d KdV). These include three inverse scattering based models---\texttt{IS-FNO$^{o}$}, \texttt{IS-FNO$^{*}$}, and \texttt{IS-FNO$'$};
three Koopman-inspired models---
\texttt{kFNO$^{o}$}, \texttt{kFNO$^{*}$}, and \texttt{kFNO$'$};
and the baseline \texttt{FNO}.
For the 1d KdV experiment, which admits an analytical IST solution, two additional restricted variants (\texttt{IS-FNO$'_{\kappa}$} and \texttt{IS-FNO$'_{\kappa3}$}) are included to assess the benefit of hard-coding known scattering structure.

For the three two-dimensional datasets, where computational cost is substantially higher, we focus on \texttt{IS-FNO$^{o}$}, \texttt{kFNO$^{*}$}, and the baseline \texttt{FNO}. The model \texttt{kFNO$^{*}$}, originally proposed in \cite{YuH2024}, serves as the strongest Koopman-based reference.

All models are trained using an $n$-step prediction loss. Specifically, all kFNO and IS-FNO variants are trained to learn the $n$-step advancement operator $\bar{G}$ by minimizing Eq.~\eqref{eq:loss_extended}, whereas the baseline \texttt{FNO} is trained to learn the single-step operator $G$ while minimizing the $n$-step recurrent loss in Eq.~\eqref{eq:loss}. For all datasets, the prediction horizon is set to $n = 20$.

For fair comparison, all 1d models use latent dimension $d_{\z} = 30$ for the baseline \texttt{FNO} and all kFNO variants, and $d_{\z_*} = d_{\z} - d_{v} = 30$ for all IS-FNO variants.  
For 2d models, due to higher computational cost, we use a reduced latent size: $d_{\z}=20$ for the baseline \texttt{FNO} and all kFNO variants, and $d_{\z_*}=d_{\z}-d_{v}=20$ for all IS-FNO variants.  
Additional architectural and training details are provided in Appendix~\ref{app:nn_detail}.

\subsection{Evaluation Metrics}

Model accuracy is assessed by direct comparison between predictions and reference numerical solutions.  
Short-term predictive accuracy is quantified using the relative $L_{2}$ error averaged over $n=20$ prediction steps, computed on the training and validation datasets using Eq.~\eqref{eq:loss} for the baseline model and Eq.~\eqref{eq:loss_extended} for the others.

Long-horizon behavior is characterized using time-accumulated errors over extended rollouts. For a $j$-step prediction this is defined as
\begin{align}
    J(j) &= \mathbb{E}_{\phi_0 \sim \chi}\!\left[ C \big( \mathcal{G}_{\hat{\theta}}^{j} \phi_0,\; G^{j} \phi_0 \big) \right], 
    &&\text{for baseline \texttt{FNO}}, \nonumber \\
    J(j) &= \mathbb{E}_{\phi_0 \sim \chi}\!\left[ C\big( (\bar{\mathcal{G}}_{\hat{\theta}}^{\lceil j/n \rceil} \phi_0)[j \bmod n],\; G^{j} \phi_0 \big) \right],
    &&\text{for other models}.
    \label{eq:long_horizon_error}
\end{align}

For chaotic or noise-sensitive stiff systems such as KS and MS, trajectory-wise agreement eventually becomes infeasible, even for the numerical solver itself. Therefore, we additionally evaluate the spatial autocorrelation
\begin{equation}
    \mathfrak{K}(r)= \lim_{T \to \infty} \mathbb{E}_{\phi_0 \sim \chi}\!\left(
    \frac{ \int_{t_*}^{t_*+T} \int_{\D} \phi(x,t)\,\phi(x-r,t)\,\text{d}x \text{d}t}
         { \int_{t_*}^{t_*+T} \int_{\D} \phi(x,t)\,\phi(x,t)  \,\text{d} x \text{d}t}
\right),
\label{eq:corr}
\end{equation}
where $\phi(x,t_*)$ denotes the model prediction after suffcient late time (e.g. $t_* \geq 1000\Delta_t$).  
This statistic measures long-range structural coherence and provides a robust diagnostic of dynamical fidelity in chaotic regimes.

Figure~\ref{fig:errTable} reports short-term (20-step) errors across all datasets (numerical values in appendix Table~\ref{Table1}).  
Long-horizon error growth is shown in Figs.~\ref{fig:ave_err_1dsiva}--\ref{fig:ave_err2d}.  
Representative trajectories and snapshots appear in Figs.~\ref{fig:KS_1D}--\ref{fig:Snapshot_2D}, with corresponding autocorrelation curves shown in Fig.~\ref{fig:corr}.

\subsection{Comparative Model Behavior Across PDE Families}

\subsubsection{\texttt{IS-FNO$^o$}: Overall Best Performing Model}

Across all non-stiff datasets, \texttt{IS-FNO$^o$} consistently attains the lowest short-term errors (Fig.~\ref{fig:errTable}, except the 4th panel) and the slowest long-horizon error growth (Figs.~\ref{fig:ave_err_1dsiva}--\ref{fig:ave_err2d}).  
Its advantages are most pronounced in the integrable and chaotic regimes.

For 1d KdV equation, \texttt{IS-FNO$^o$} accurately reproduces long-horizon dynamics for both initialization types---soliton superposition \eqref{eq:init_1dsol} and low-wave-number randomization \eqref{eq:init_lowwavenumber}.  
As shown in Fig.~\ref{fig:KdV_1D}, 2000-step rollouts are nearly indistinguishable from reference solutions, with accumulated errors remaining below $2\%$ (Fig.~\ref{fig:ave_err_1dkdv}).  
A similar level of fidelity is observed for the 2d KdV/KP equation, as demonstrated by the bottom two rows of Fig.~\ref{fig:Snapshot_2D} and the third panel of Fig.~\ref{fig:ave_err2d}.

\begin{figure*}
    \centerline{
        \includegraphics[width=1\linewidth]{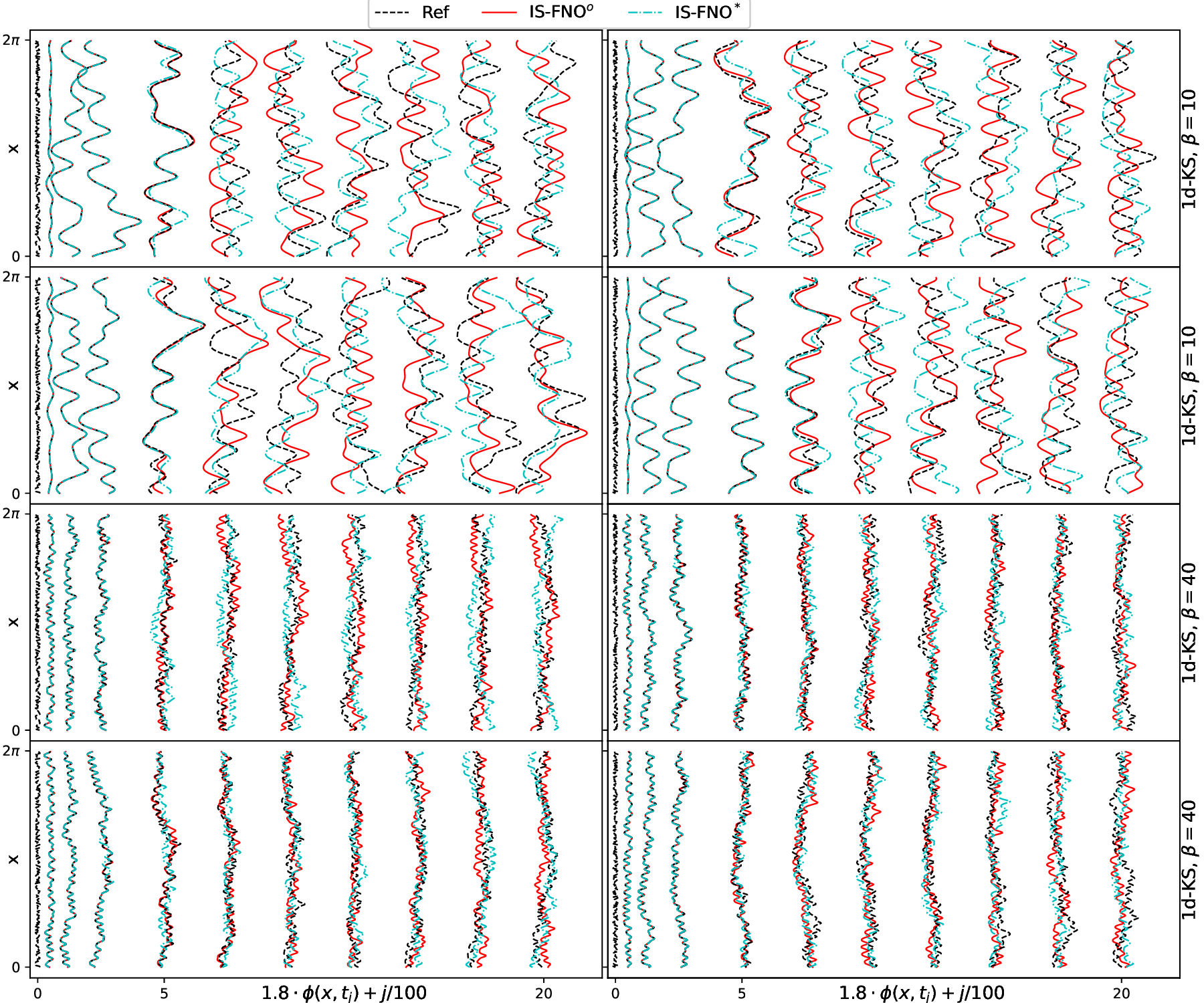} 
    }
    \caption{
        \label{fig:KS_1D}
        Long-horizon solution trajectories $\phi(x,t_j)$ of the 1d-KS equation \eqref{eq:KS} at $\beta = 10$ (top two rows) and $\beta = 40$ (bottom two rows).  
        Black curves denote reference solutions computed using a spectral solver. Predictions from \texttt{IS-FNO$^o$} (red) and \texttt{IS-FNO$^{*}$} (cyan) are shown for comparison.  
        Each panel contains four randomly initialized sequences, with eleven temporal snapshots at $t_j = j\,\Delta_t$ for  
        $j=\{0,\,50,\,125,\,250,\,500,\,750,\,1000,\,1250,\,1500\,1750,2000\}$.  
        For visualization, each snapshot $\phi(x,t_j)$ is amplified by a factor of $1.8$ and shifted by $j/100$ to prevent overlap.
    }
\end{figure*}

For chaotic KS equations at different $\beta$, the model provides accurate short-term predictions and stable long-horizon rollouts (Figs.~\ref{fig:KS_1D}, \ref{fig:Snapshot_2D}), and it matches long-term statistical behavior (Fig.\ref{fig:corr}).  

In MS equations at moderate $\beta$ (10 and 15), \texttt{IS-FNO$^o$} captures the formation of giant cusp front in 1d (Fig.~\ref{fig:MS_1D}) and cellular flame fronts in 2d (Fig.~\ref{fig:Snapshot_2D}).  While noise perturbations appears during later model rollouts, autocorrelation statistics remain accurate (Fig.~\ref{fig:corr}).

\subsubsection{Nonlinear Exponential Fourier Layers vs.\ Vanilla Stacked Fourier Layers}

Within the IS-FNO architecture, parameterizing the latent operator $A$ using the nonlinear exponential Fourier layer (\ref{eq:ExpFourierLayer}) provides a clear and consistent advantage over using vanilla stacked Fourier layers (\ref{eq:FourierLayer}).

This trend is most evident when comparing \texttt{IS-FNO$^o$} with \texttt{IS-FNO$^{*}$} across all four non-stiff 1d datasets (KS at two $\beta$ values, MS at $\beta=10$, and KdV).  
\texttt{IS-FNO$^o$} systematically achieves lower training and validation errors (red vs.\ purple bars in Fig.~\ref{fig:errTable}) and lower rollout errors (red-solid vs.\ red-dashed curves in Figs.~\ref{fig:ave_err_1dsiva} and \ref{fig:ave_err_1dkdv}).  
The advantage is visible not only in short-horizon prediction accuracy but also in reduced long-term error accumulation.

A similar but weaker trend appears in the Koopman-based architecture.  
When comparing \texttt{kFNO$^o$} to \texttt{kFNO$^{*}$}, the former again shows lower short-term errors for all four non-stiff 1d datasets (green vs.\ blue bars in Fig.~\ref{fig:errTable}).  
Rollout errors are also slightly lower for \texttt{kFNO$^o$} in the two KS cases (blue-dashed vs.\ blue-solid curves in Fig.~\ref{fig:ave_err_1dsiva}).  
However, the long-term behavior is less consistent: in the KdV dataset (Fig.~\ref{fig:ave_err_1dkdv}), the simpler \texttt{kFNO$^{*}$} becomes more accurate over very long rollouts, reversing the short-term advantage of the nonlinear exponential layer.

Overall, these comparisons indicate that the nonlinear exponential Fourier layer reliably improves latent dynamics modeling within IS-FNO, while its benefit in the Koopman-based setting is more equation-dependent and may diminish over very long integration horizons.

\subsubsection{Nonlinear vs.\ Linear Exponential Fourier Layers to learn Latent Dynamics.}

Within the IS-FNO architecture, where the latent operator $A$ is represented by an exponential Fourier layer, enabling the nonlinear term ($\gamma = 1$ in Eq.~\ref{eq:ExpFourierLayer}) provides a clear performance advantage over its linear counterpart ($\gamma = 0$).  
Across all five 1d datasets, \texttt{IS-FNO$^o$} consistently outperforms the linear variant \texttt{IS-FNO$'$}, as seen from the training/validation errors (red vs.\ yellow bars in Fig.~\ref{fig:errTable}) and from the rollout behavior (red-solid vs.\ red-dotted curves in Figs.~\ref{fig:ave_err_1dsiva} and \ref{fig:ave_err_1dkdv}).

A similar pattern appears in the Koopman-inspired architecture.  
The nonlinear model \texttt{kFNO$^o$} generally improves upon \texttt{kFNO$'$} (green vs.\ cyan bars in Fig.~\ref{fig:errTable}; blue-solid vs.\ blue-dotted curves in Fig.~\ref{fig:ave_err_1dsiva}).  
The primary exception is the 1d KdV dataset, where the linear variant eventually overtakes the nonlinear one over very long rollouts.  
This behavior is consistent with the predominantly linear structure of the KdV scattering dynamics (Eq.~\ref{eq:kdv-scatter}),
together with the freedom that Koopman lift-project maps are not constrained to be exactly reversible, 
making the linear latent evolution particularly well suited for this integrable system.

\subsubsection{Inverse-Scattering Architecture vs.\ Koopman Architectures}

\begin{figure*}
    \centering
    \includegraphics[width=1\linewidth]{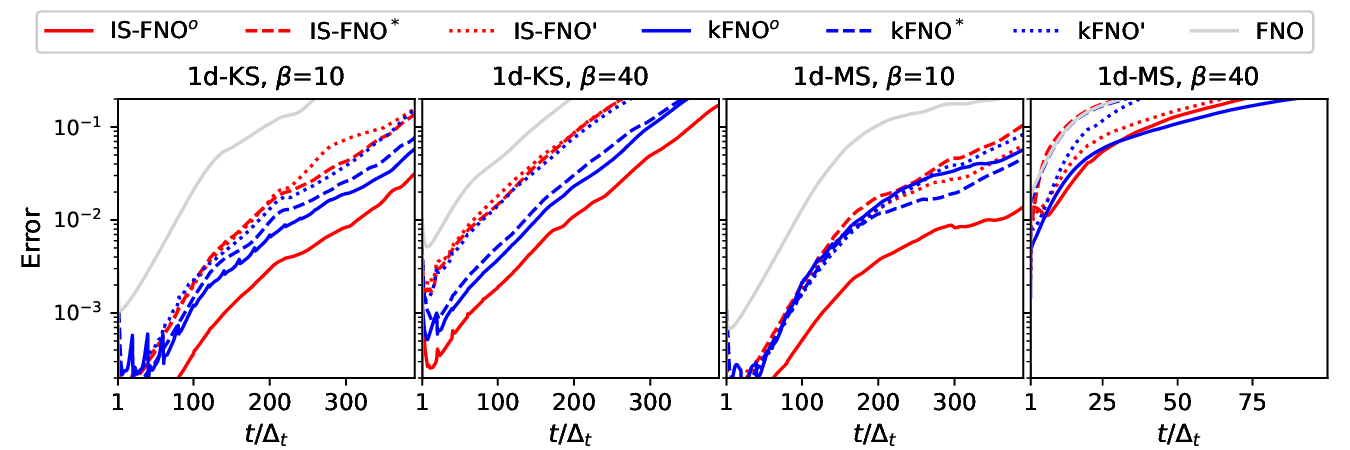}
    \caption{
        \label{fig:ave_err_1dsiva}
        Long-horizon error evolution \eqref{eq:long_horizon_error} for seven models evaluated on four 1d datasets (KS and MS at $\beta = 10$ and $\beta = 40$).  
        For each model and dataset, errors are averaged over 20 rollout sequences by uniform random initialization, Eq.~\eqref{eq:MS_KS_init}.
    }
\end{figure*}

When the latent operator $A$ is modeled using nonlinear exponential Fourier layers, the inverse scattering architecture provides a clear advantage over the Koopman-inspired architecture.  
Comparing \texttt{IS-FNO$^o$} with \texttt{kFNO$^o$} in learning the four non-stiff 1d datasets, the former consistently yields lower short-term errors
(red vs.\ green bars in Fig.~\ref{fig:errTable})
 and reduced long-horizon error accumulation
 (red-solid vs.\ blue-solid curves in the first three panels of Fig.~\ref{fig:ave_err_1dsiva} and in Fig.~\ref{fig:ave_err_1dkdv}).
This advantage stems from the near-reversible lifting-projection pairing in the inverse scattering design, which stabilizes nonlinear latent-time evolution.

When both architectures are restricted to \emph{linear} exponential layers (\texttt{IS-FNO$'$} vs.\ \texttt{kFNO$'$}), their performance becomes nearly indistinguishable
(cyan vs.\ yellow bars in Fig.~\ref{fig:errTable}; blue vs.\ red dotted curves in Figs.~\ref{fig:ave_err_1dsiva} and \ref{fig:ave_err_1dkdv}), 
indicating that the performance gap between the two architectures is primarily attributed to the treatment of nonlinear latent dynamics, where the reversibility constraint yields the greatest benefit.

\subsubsection{Performance on the Stiff MS Equation at $\beta=40$}

The stiff MS dataset poses difficulties for all Fourier-based models due to extreme sensitivity to numerical noise and limited dataset resolution (recall from Sec.~\ref{sec:1ddata} that the reference spectral solver becomes unstable by $\beta=50$). All models show signs of overfitting, reflected by large discrepancies between training and validation errors (fourth panel in Fig.~\ref{fig:errTable}).  

\begin{figure*}
    \centerline{
        \includegraphics[width=1\linewidth]{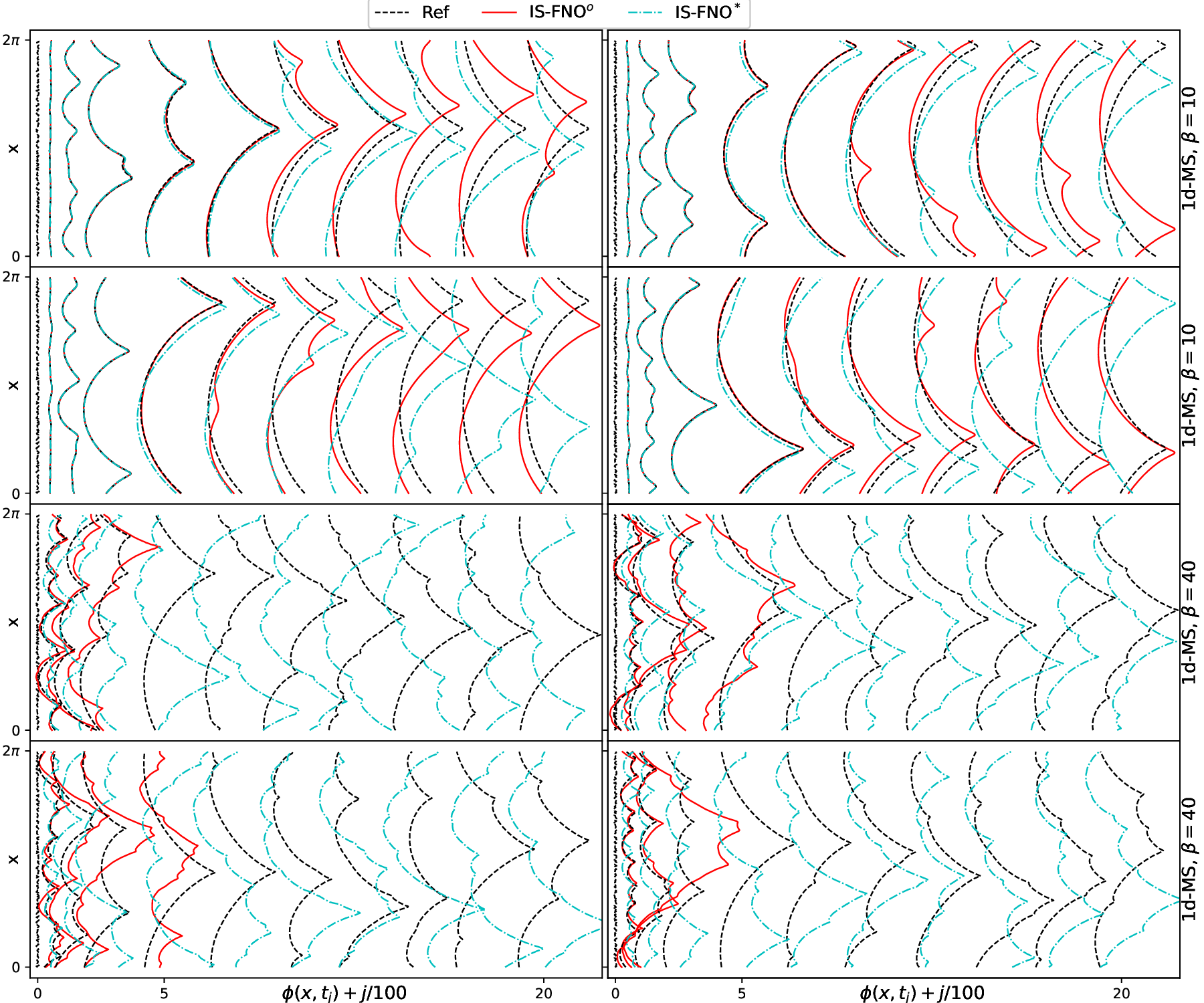} 
    }
    \caption{
        \label{fig:MS_1D}
        Long-term reference solutions of the 1d-MS equation \eqref{eq:MS} at $\beta = 10$ and $\beta = 40$, compared against predictions from \texttt{IS-FNO$^o$} and \texttt{IS-FNO$^{*}$}.  
        Multiple randomly initialized trajectories are included to assess long-horizon predictive stability.  
        Note that predictions from \texttt{IS-FNO$^o$} diverge after a few hundred time steps for the stiff case at $\beta = 40$.
    }
\end{figure*}

Among the seven models evaluated, \texttt{IS-FNO$^o$} achieves relatively low short-term error but diverges during long rollouts (Fig.~\ref{fig:MS_1D}).  
By contrast, \texttt{IS-FNO$^{*}$} and \texttt{kFNO$^{*}$} maintain stable long-horizon trajectories despite higher short-term errors. Both replicate the correct autocorrelation statistics (Fig.~\ref{fig:corr}).

These findings indicate that vanilla stacked Fourier layers, while less accuate in learning short time prediction, provide improved robustness in extremely stiff regimes where error is dominated by small-scale noise amplification.

\subsubsection{Performance on the 1d KdV Equation}

\begin{figure*}
    \centering
    \includegraphics[width=1\linewidth]{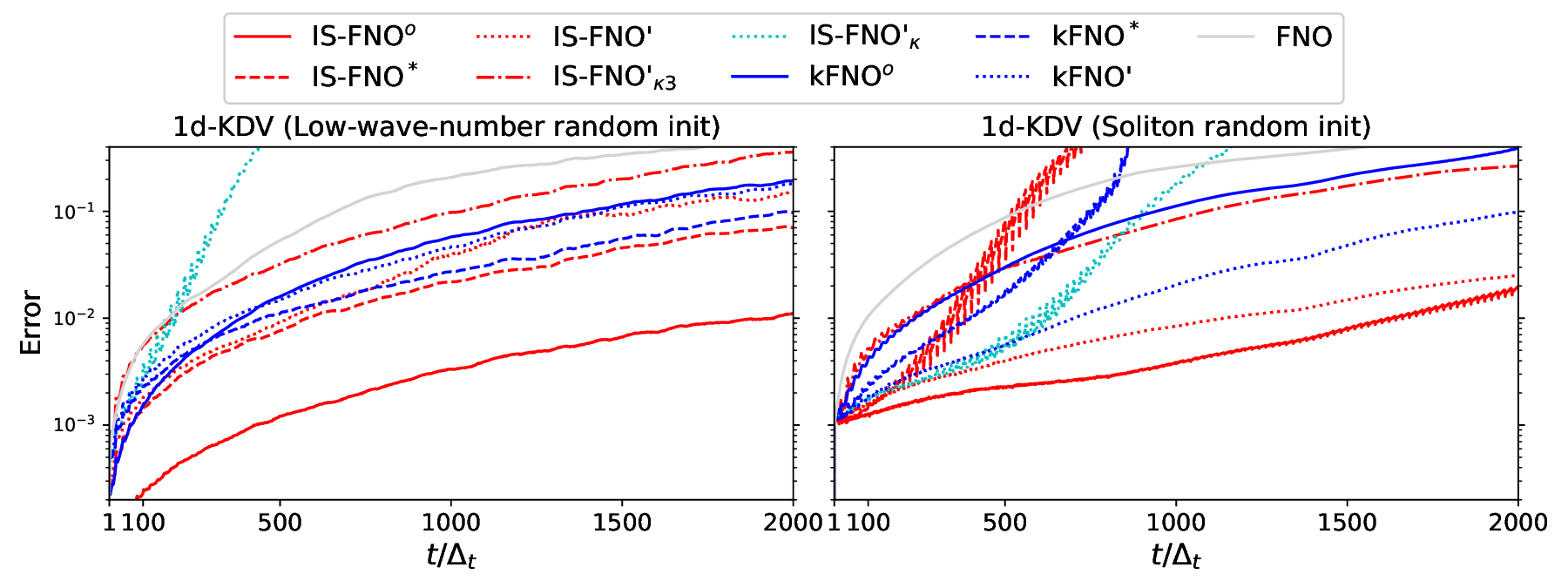}
    \caption{
        \label{fig:ave_err_1dkdv}
        Long-horizon error evolution for nine models trained on the 1d-KdV dataset.  
        Error is averaged over 20 rollout sequences initialized by low-wavenumber randomization (Eq.~\ref{eq:init_lowwavenumber}, left ) and random soliton superposition (Eq.~\ref{eq:init_1dsol}, right).
    }
\end{figure*}

The integrable KdV dataset is substantially easier to learn than the stiff MS dataset or the chaotic KS dataset.  
All full-capacity models using exponential Fourier layers---\texttt{IS-FNO$^o$}, \texttt{IS-FNO$'$}, \texttt{kFNO$^o$}, and \texttt{kFNO$'$}---achieve low accumulated error over 2000-step long-horizon rollouts (below $40\%$ for both initialization types).  
These behaviors correspond to the red-solid, red-dotted, blue-solid, and blue-dotted curves in the two panels of Fig.~\ref{fig:ave_err_1dkdv}.

Models using stacked vanilla Fourier layers (\texttt{IS-FNO$^{*}$}, \texttt{kFNO$^{*}$}) perform well for low-wavenumber initializations but degrade substantially for multi-soliton initial conditions, highlighting the difficulty of capturing soliton interactions with purely vanilla Fourier layers.

The two reduced inverse scattering variants, \texttt{IS-FNO$'_{\kappa}$} and \texttt{IS-FNO$'_{\kappa3}$}, unsurprisingly exhibit higher short-term errors due to their restricted parameterization of the latent scattering evolution (orange and violet bars in Fig.~\ref{fig:errTable}).  
They also show larger long-horizon rollout errors (cyan-dotted and red-dash-dotted curves in Fig.~\ref{fig:ave_err_1dkdv}).

Interestingly, \texttt{IS-FNO$'_{\kappa3}$}, which hardcodes the analytical $\kappa^{3}$ KdV scattering scaling \eqref{eq:kdv-scatter}, achieves better long-term stability than \texttt{IS-FNO$'_{\kappa}$} despite having the largest short-term error among all models.  
After 2000 rollout steps, \texttt{IS-FNO$'_{\kappa3}$} reaches an error level below $40\%$ for both initialization types, comparable to the performance of the full-capacity \texttt{kFNO$^o$} model.

\begin{figure*}
    \centerline{
        \includegraphics[width=1\linewidth]{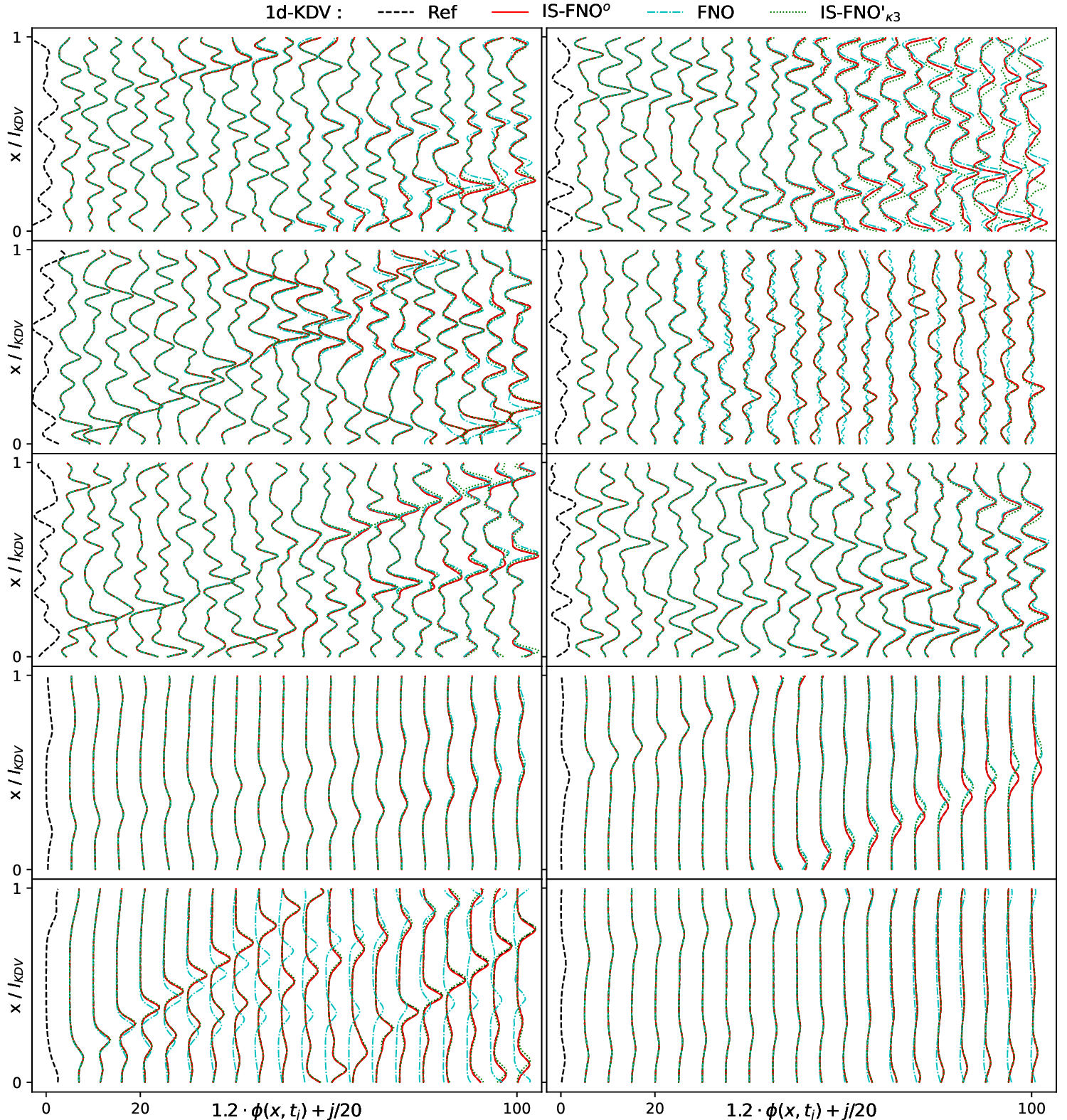}
    }
    \caption{
        \label{fig:KdV_1D}
        Long-term solution sequences $\phi(x,t_j)$ of the 1d-KdV equation (Eq.~\eqref{eq:kdv}), shown at times $t_j = j \cdot 100\,\Delta_t$ for $j = 0,\ldots,20$.  
        Reference solutions from the spectral solver (black) are compared with predictions from \texttt{IS-FNO$^o$} (red), \texttt{FNO} (cyan), and \texttt{IS-FNO$'_{\kappa3}$} (green).  
        The top three rows (six sequences) correspond to initial conditions generated via low-wavenumber randomization (Eq.~\ref{eq:init_lowwavenumber}), while the bottom two rows (four sequences) correspond to initial conditions constructed from random soliton superpositions (Eq.~\ref{eq:init_1dsol}).
    }
\end{figure*}

Furthermore, \texttt{IS-FNO$'_{\kappa3}$} outperforms the baseline \texttt{FNO} in long-horizon predictions across ten randomly sampled rollout sequences.  
As shown in Fig.~\ref{fig:KdV_1D}, the cyan curves (representing \texttt{IS-FNO$'_{\kappa3}$}) more closely track the reference solution (black curves) than the baseline \texttt{FNO} (green curves), exhibiting fewer noticeable deviations.  
These results highlight the value of embedding analytical structure into reduced models, even when expressive capacity is limited.

\subsubsection{Two-Dimensional Dynamics}

\begin{figure*}
    \centerline{
        \includegraphics[width=1\linewidth]{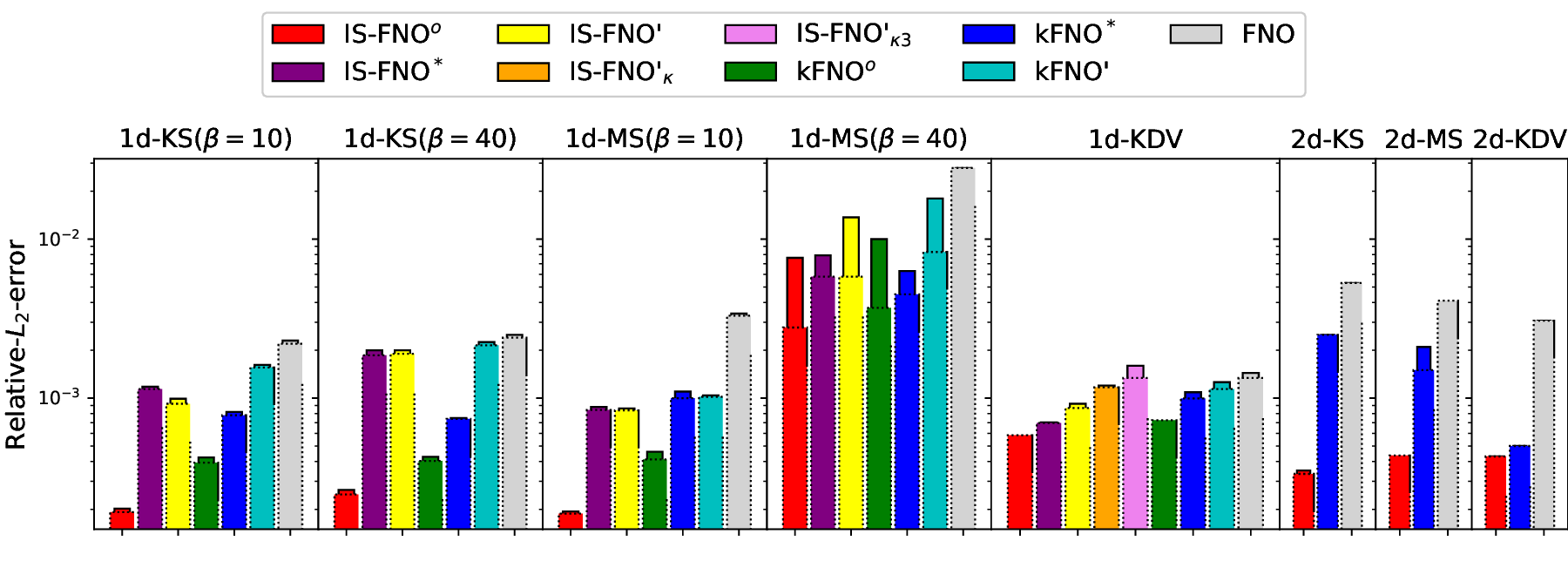}
    }
    \caption{
        \label{fig:errTable}
        Short-term relative $L_2$ errors for training (wide bars, dotted outlines) and validation (thin bars, solid outlines) across eight datasets: five 1d cases (KS and MS at $\beta=10$ and $\beta=40$, and KdV) and three 2d cases (KS and MS at $\beta=15$, and KDV/KP). Numerical values are listed in Table~\ref{Table1} in the Appendix.
    }
\end{figure*}

\begin{figure*}
    \centering
    \includegraphics[width=0.8\linewidth]{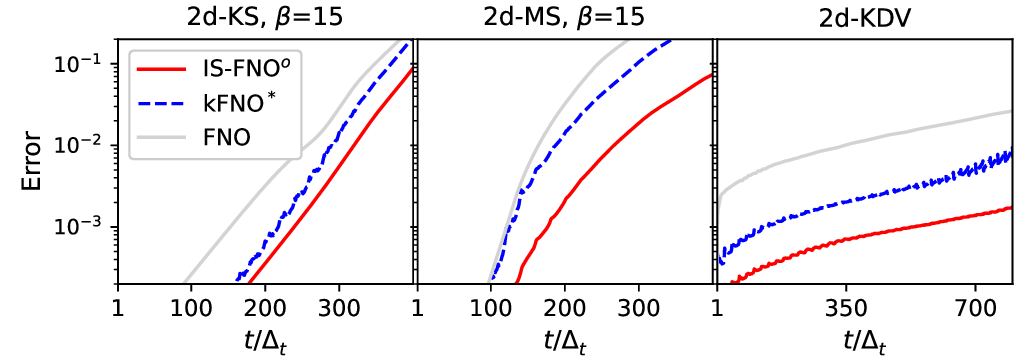}
    \caption{
        \label{fig:ave_err2d}
        Long-horizon error evolution \eqref{eq:long_horizon_error} for three models trained on three 2d datasets.  
        Errors are averaged over five rollout sequences initialized by low-wavenumber randomization.
    }
\end{figure*}

Results in two dimensions mirror those in one dimension.  
Across all three 2d datasets, \texttt{IS-FNO$^o$} consistently delivers the lowest short-term errors (rightmost panels of Fig.~\ref{fig:errTable}) and the slowest rollout error accumulation (Fig.~\ref{fig:ave_err2d}), outperforming both \texttt{kFNO$^{*}$} and baseline \texttt{FNO}.

\begin{figure*}
    \centering
    \includegraphics[width=0.9\linewidth]{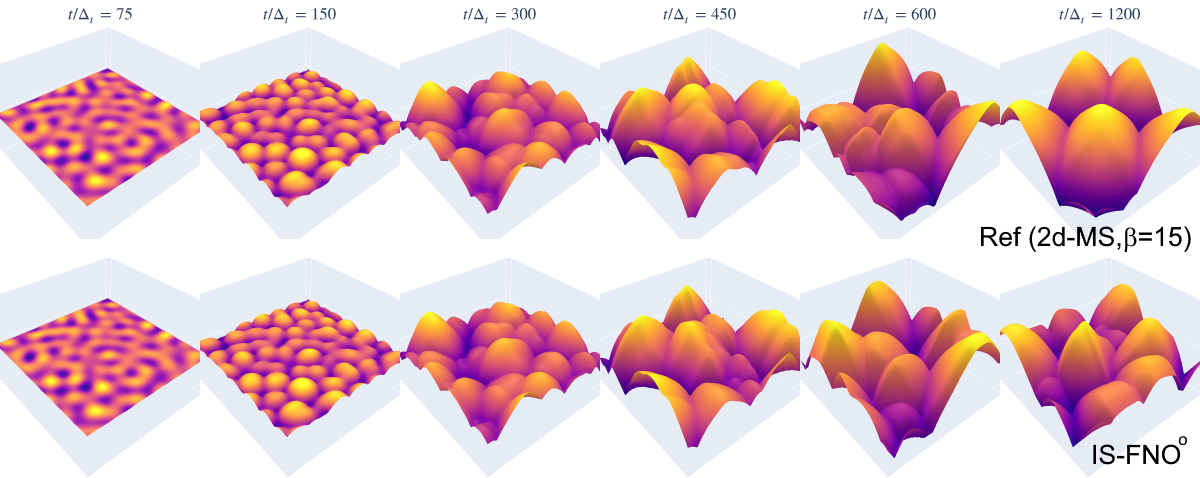}
    \includegraphics[width=0.9\linewidth]{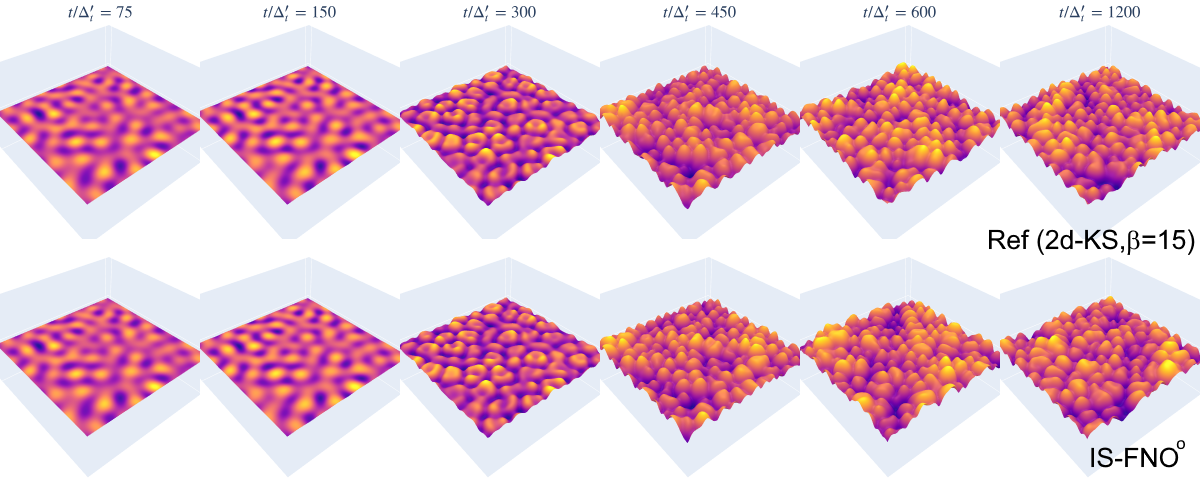}
    \includegraphics[width=0.9\linewidth]{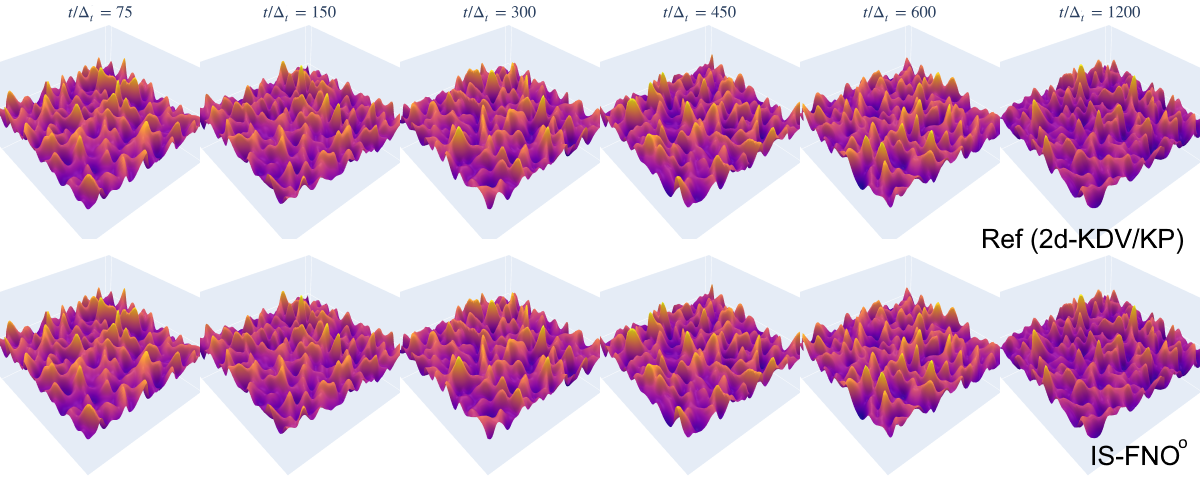} 
    \caption{
        \label{fig:Snapshot_2D}
        Comparison of randomly selected solution snapshots $\phi(x,t_j)$ at times $t_j = j \Delta_t$ for $j=\{75,150,300,450,600,1200\}$.  
        Results are shown for the 2d-MS equation (top two rows) and the 2d-KS equation (middle two rows), both at $\beta = 15$, as well as for the 2d-KdV/KP equation (bottom two rows).  
        Each sequence includes the reference solutions obtained from the spectral solver and the corresponding predictions produced by \texttt{IS-FNO$^o$}.
    }
\end{figure*}

\begin{figure*}
    \centerline{
        \includegraphics[width=0.65\linewidth]{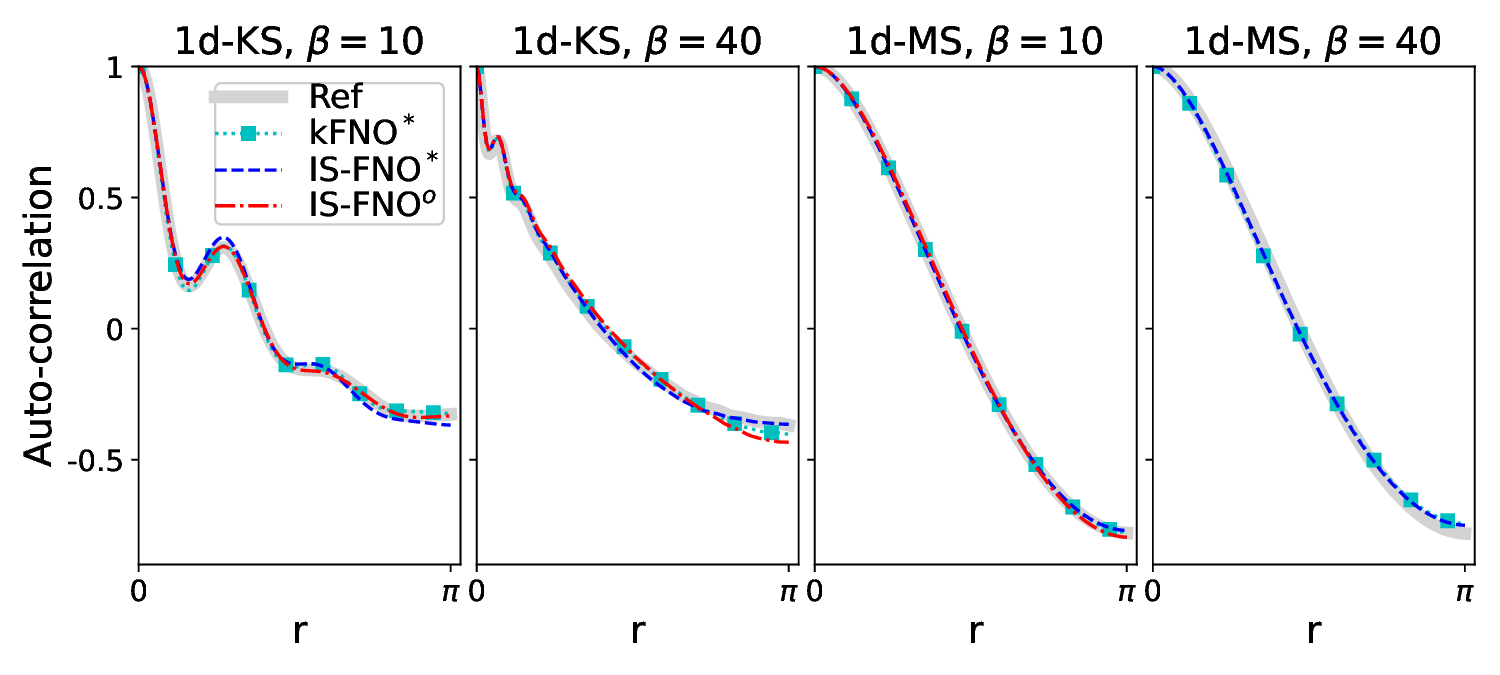}
        \includegraphics[width=0.335\linewidth]{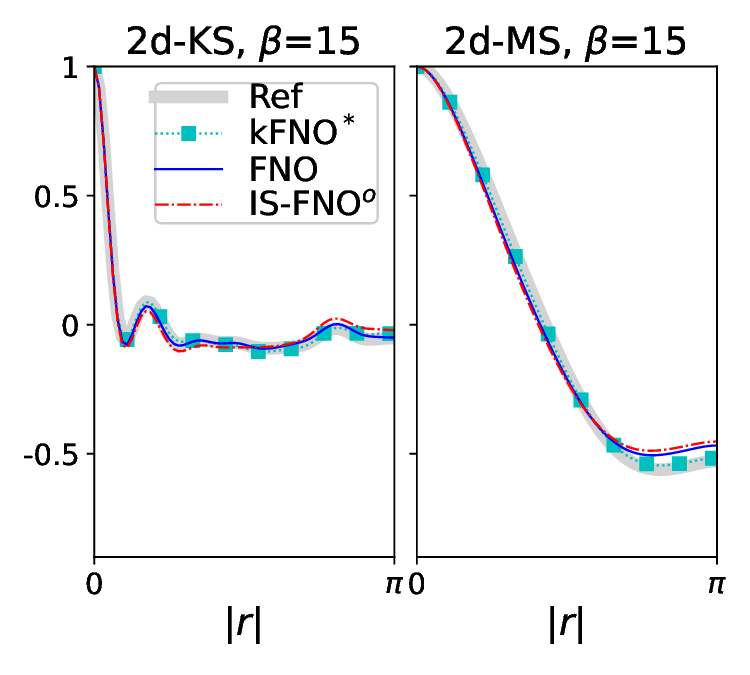}
    }
    \caption{
        \label{fig:corr}
Autocorrelation function (Eq.~\ref{eq:corr}) comparing long-term statistics of reference solutions and model predictions.
Columns 1--4 show 1d KS and MS equations at $\beta=10$ and $40$; columns 5--6 show 2d KS and MS at $\beta=15$.
Predictions are from \texttt{kFNO$^{*}$}, baseline \texttt{FNO}, \texttt{IS-FNO$^{*}$}, and \texttt{IS-FNO$^o$}.
Statistics are averaged over $>10$ randomly intialized trajectories for $1000 < t/\Delta_t < 4000$.
    }
\end{figure*}

Snapshots in Fig.~\ref{fig:Snapshot_2D} show that \texttt{IS-FNO$^o$} captures complex nonlinear structures and fine-scale patterns with high accuracy.  
Autocorrelation statistics confirm that the model also reproduces correct long-term spatial coherence (Fig.~\ref{fig:corr}).

\section{Summary and Conclusion}
\label{sec:conclusion}

This work proposed an inverse scattering inspired neural operator framework for learning solution time-advancement operators of nonlinear partial differential equations.
The design is motivated by the structural insight of the classical inverse scattering transform (IST), which decomposes nonlinear evolution into a forward scattering map, a simple evolution in spectral space, and an inverse scattering reconstruction.
Adapting this idea to the neural-operator setting, the proposed IS-FNO enforces a near-reversible pairing between lifting and projection maps and employs exponential Fourier layers to model latent temporal evolution in a structured and physically motivated manner.

A comprehensive set of numerical experiments was conducted on eight benchmark datasets spanning integrable, chaotic, stiff, and noise-sensitive regimes, including one- and two-dimensional variants of the Michelson--Sivashinsky, Kuramoto--Sivashinsky, Korteweg--de Vries, and Kadomtsev--Petviashvili equations.
Across these datasets, IS-FNO consistently outperformed both the baseline FNO and Koopman-inspired FNO architectures in terms of short-term accuracy and long-horizon stability, particularly when nonlinear exponential Fourier layers were used to parameterize latent dynamics.

Several key observations emerge from the results.
First, enforcing near-reversibility between lifting and projection maps significantly stabilizes nonlinear latent-time evolution, leading to reduced error accumulation during rollout.
This advantage is most pronounced when combined with nonlinear exponential Fourier layers, which outperform both linear exponential layers and stacked vanilla Fourier layers in non-stiff regimes.
Second, the benefit of nonlinear latent evolution is equation dependent: for integrable systems such as the KdV equation, linear exponential dynamics already capture the dominant scattering behavior, while for chaotic systems, nonlinear latent interactions are essential for accurate long-term modeling.
Third, in extremely stiff and noise-sensitive regimes, such as the Michelson--Sivashinsky equation at large $\beta$, simpler vanilla Fourier layers can offer improved robustness at the expense of short-term accuracy, underscoring an inherent trade-off between expressiveness and numerical stability.

The study also demonstrates that embedding analytical structure into reduced models can be highly effective.
In particular, a restricted IS-FNO variant that hard-codes the $\kappa^3$ scattering scaling of the KdV equation achieves competitive long-horizon performance despite limited parameterization, outperforming unconstrained baseline models in long-term predictions.

Overall, this work highlights the importance of architectural inductive biases---specifically reversibility and physically motivated spectral evolution---in neural operator design.
The proposed IS-FNO framework provides a flexible yet principled approach for learning time-advancement operators of nonlinear PDEs, offering improved stability, interpretability, and generalization across a broad range of dynamical regimes.

Possible futher works include refining the architecture to better align with classical scattering structure, where integrable systems exhibit continuous spectrum (radiation) and discrete eigenvalues (solitons); disentangling these components could improve interpretability and accuracy.
Extending IS-FNO to continuous-time modeling via adjoint methods as in neural ODEs \cite{chen_NeuralODE} is another avenue, offering adaptive time-stepping and improved handling of irregularly sampled data.
Additional directions include improved treatment of diverse boundary conditions and noisy data, applications to more complex multi-physics systems, adaptive spectral parameterizations, and tighter integration of analytical structure for partially integrable or weakly chaotic dynamics.

\begin{acknowledgments}
The author gratefully acknowledges the financial support by the Swedish Research Council(VR-2019-05648) and the AI Lund initiative grant.
The computations were enabled by resources provided by the National Academic Infrastructure for Supercomputing in Sweden (NAISS), at ALVIS and Tetralith, partially funded by the Swedish Research Council through grant agreement no. 2022-06725.
\end{acknowledgments}

\section*{Data Availability Statement}
The code and data that support the findings of this study will be openly available at www.github.com/RixinYu/ML\_paraFlame upon acceptance for publication.

\section*{Declaration of Interest}
The author has no conflicts to disclose.



\appendix

\section{Inverse Scattering Transform of 1d KdV Equation \label{app:ist}}

The one-dimensional KdV equation \eqref{eq:kdv} can be recast as a \textit{Lax equation}
\begin{equation}
\partial_t \mathcal{L} = [\mathcal{M}, \mathcal{L}] = \mathcal{M}\mathcal{L} - \mathcal{L}\mathcal{M},
\label{eq:Lax}
\end{equation}
for a pair of differential operators $(\mathcal{L}, \mathcal{M})$ defined as
\begin{equation}
\mathcal{L} := -\partial_x^2 + \phi(x,t), 
\qquad 
\mathcal{M} := -4\partial_x^3 + 6 \phi \partial_x + 3 \phi_x.
\end{equation}
Here, $\mathcal{L}$ is a Schrödinger-type operator with potential $\phi(x,t)$. The corresponding eigenvalue problem is
\begin{equation}
\mathcal{L}\psi = -\psi_{xx} + \phi(x,t)\psi = \lambda \psi,
\label{eq:schr}
\end{equation}
where $\psi$ is an eigenfunction associated with eigenvalue $\lambda$. 
From the Lax equation \eqref{eq:Lax}, it follows that $\mathcal{L}$ and $\mathcal{M}$ evolve compatibly such that the eigenvalues $\lambda$ are time-invariant (the \textit{isospectral property}), while the eigenfunctions evolve according to
\begin{equation}
\partial_t \psi = -\mathcal{M}\psi.
\label{eq:eigM}
\end{equation}

The inverse scattering transform (IST) method\cite{IST_gardner1967method,IST_ablowitz1981solitons,IST_novikov1984theory} for Eq. \eqref{eq:kdv} consists of three conceptual steps corresponding to Eqs.~\ref{eq:IST_forward}-\ref{eq:IST_inverse} in the introduction section:

1.\textit{Forward scattering transform}:
Given an initial condition $\phi(x,0)$, one solves the stationary Schrödinger problem \eqref{eq:schr} to determine the scattering data
\begin{equation}
\mathcal{S}(0) = \{\, T(k), R^{\pm}(k,0);\, c^{\pm}_1(0),\dots,c^{\pm}_N(0) \,\},
\end{equation}
where $T(k)$ is the transmission coefficient, $R^{\pm}(k,t)$ are the left and right reflection coefficients corresponding to the continuous spectrum, and $c_j^{\pm}(t)$ are the norming constants associated with discrete eigenvalues $\lambda_j = -k_j^2$.

2.\textit{ Time evolution of scattering data}:  
The time evolution of these coefficients follows directly from \eqref{eq:eigM} as
\begin{align}
R^{-}(k,t) &= R^{-}(k,0)\,e^{-8i k^3 t}, \nonumber \\
R^{+}(k,t) &= R^{+}(k,0)\,e^{+8i k^3 t}, \nonumber \\
c^{-}_j(t) &= c^{-}_j(0)\,e^{+4 k_j^3 t}, \quad j = 1,\dots,N, \nonumber \\
c^{+}_j(t) &= c^{+}_j(0)\,e^{-4 k_j^3 t}, \quad j = 1,\dots,N,
\label{eq:kdv-scatter}
\end{align}
while the transmission coefficient $T(k)$ and eigenvalues $k_j$ remain constant in time.

3. \textit{Inverse scattering transform}:
The field $\phi(x,t)$ is reconstructed from $\mathcal{S}(t)$ by solving the Gelfand--Levitan--Marchenko integral equation \cite{GLM_gel1951determination,GLM_marchenko2011sturm}
\begin{equation}
K(x,z,t) + F(x+z,t) + \int_x^{\infty} K(x,y,t)F(y+z,t)\, \mathrm{d}y = 0,
\label{eq:GLM}
\end{equation}
where
\begin{equation}
\phi(x,t) = -2\,\partial_x K(x,x,t),
\end{equation}
and the kernel $F(x,t)$ is defined in terms of the scattering data as
\begin{equation}
F(x,t) = \frac{1}{2\pi} \int_{-\infty}^{\infty} R^{+}(k,t)e^{ikx}\, \mathrm{d}k
+ \sum_{j=1}^{N} c^{-}_j(t)^{\,2}\, e^{-k_j x}.
\end{equation}

The IST method thus maps the nonlinear PDE \eqref{eq:kdv} into a linear evolution in the scattering space. 
For reflectionless potentials ($R^{\pm}\equiv 0$), only discrete eigenvalues contribute, leading to $N$-soliton solutions. 
The entire procedure can be viewed as a nonlinear generalization of the Fourier transform for integrable PDEs: 
the forward transform converts $\phi(x,0)$ into scattering coefficients, which evolve linearly in time, 
and the inverse transform reconstructs $\phi(x,t)$ from the evolved scattering data.

\section{ Model hyper-parameters and training details \label{app:nn_detail} }

All IS-FNO models employ the zero-stacking lifting map $L_0$ and set $d_{\z_a} = d_v$, which avoids the final $f$-inversion step and therefore reduces computational cost.
To learn the four 1d MS and KS datasets, which contain significant high-frequency content, all 1d models use Fourier layers (either vanilla or exponential) with $\kappa^{\max} = 128$.  
For the 1d KdV dataset, which is less demanding, we set $\kappa^{\max} = 32$ to allow truncation of higher modes.  
To learn the three 2d datasets, all models use Fourier layers with $\kappa^{\max}_{1} = \kappa^{\max}_{2} = 64$.
In all kFNO models, the projection map $P$ is implemented as a two-layer perceptron with a hidden width of 128. In all models, the nonlinear activation function $\sigma$ is chosen to be GELU.

All models are trained for 1000 epochs using the Adam optimizer with learning rate 0.0025, weight decay 1e-6 and epsilon value 1e-6. A learning rate scheduler with step size 100 and decay factor 0.5 is applied. Gradient clipping with maximum norm 10 ensures training stability.

All training is performed on a single GPU (NVIDIA A40).
For 1d models, training uses batch size 512. Training times on the 1d KdV dataset vary by architecture: \texttt{kFNO$'$} (41 min), \texttt{kFNO$^o$} (49 min), \texttt{kFNO$^*$} (54 min), \texttt{IS-FNO$'$} (57 min), \texttt{IS-FNO$^o$} (65 min), \texttt{IS-FNO$^*$} (71 min), and baseline \texttt{FNO} (94 min).
On the 1d-KS dataset ($\beta=40$), training times are: \texttt{kFNO$'$} (70 min), \texttt{kFNO$^o$} (93 min), \texttt{kFNO$^*$} (95 min), \texttt{IS-FNO$'$} (107 min), \texttt{IS-FNO$^o$} (131 min), \texttt{IS-FNO$^*$} (125 min), and baseline \texttt{FNO} (168 min).
For 2d model, training uses batch size 32. For 2d KDV/KP dataset, training time are: \texttt{kFNO$^{*}$} (46 hours), \texttt{IS-FNO$^o$} (74 hour) and baseline \texttt{FNO$^{*}$} (94 hours)

\begin{table}[h!]
\centering
\caption{Short-term relative $L_2$ training / validation errors for all models. \label{Table1}  }
\begin{tabular}{|l|c|c|c|c|c|c|c|c|c|c}
\hline
Model 
& 1d-MS($\beta$=40) 
& 1d-MS($\beta$=10)
& 1d-KS($\beta$=40) 
& 1d-KS($\beta$=10)
& 1d-KDV 
\\
\hline

\texttt{IS-FNO$^o$}
& 2.8e-3/7.6e-3
& 1.9e-4/1.9e-4
& 2.5e-4/2.7e-4
& 1.9e-4/2.0e-4
& 5.9e-4/5.3e-4
\\

\texttt{IS-FNO$'$}
& 5.8e-3/1.4e-2
& 8.4e-4/8.6e-4
& 1.9e-3/2.0e-3
& 9.2e-4/9.9e-4
& 8.7e-4/9.2e-4
\\

\texttt{IS-FNO$^*$}
& 5.8e-3/7.9e-3
& 8.4e-4/8.8e-4
& 1.9e-3/2.0e-3
& 1.1e-3/1.2e-3
& 7.0e-4/7.0e-4
\\

\texttt{kFNO$^o$}
& 3.7e-3/1.0e-2
& 4.1e-4/4.6e-4
& 4.0e-4/4.3e-4
& 3.9e-4/4.2e-4
& 7.3e-4/7.1e-4
\\

\texttt{kFNO$'$}
& 8.3e-3/1.8e-2
& 1.0e-3/1.0e-3
& 2.2e-3/2.3e-3
& 1.6e-3/1.6e-3
& 1.1e-3/1.3e-3
\\

\texttt{kFNO$^*$}
& 4.5e-3/6.3e-3
& 1.0e-3/1.1e-3
& 7.4e-4/7.5e-4
& 7.8e-4/8.2e-4
& 1.0e-4/1.1e-3
\\

\texttt{FNO}
& 2.8e-2/2.8e-2
& 3.3e-3/3.4e-3
& 2.4e-3/2.5e-3
& 2.2e-3/2.3e-3
& 1.3e-3/1.4e-3
\\

\texttt{IS-FNO$'_\kappa$}
& --
& --
& --
& --
& 1.2e-3/1.2e-3
\\

\texttt{IS-FNO$'_{\kappa3}$}
& --
& --
& --
& --
& 1.3e-3/1.6e-3
\\
\hline
\end{tabular}

\begin{tabular}{|l|c|c|c|}
\hline
Model & 2d-MS($\beta=15$) & 2d-KS($\beta=15$) & 2d-KDV/PV \\
\hline

\texttt{IS-FNO$^o$}
& 4.4e-4/3.2e-4
& 3.3e-4/3.5e-4
& 4.3e-4/4.3e-4
\\

\texttt{kFNO$^*$}
& 1.5e-3/2.1e-3
& 2.5e-3/2.5e-3
& 5.0e-4/5.0e-4
\\

\texttt{FNO}
& 4.1e-3/3.0e-3
& 5.3e-3/5.3e-3
& 3.1e-3/3.1e-3
\\

\hline
\end{tabular}

\end{table}

\bibliography{IS-FNO}

@article{Yu_Koop2025_2,
  title={Koopman-Inspired Operator Learning for Intrinsic Flame Instabilities},
  author={R. Yu and M. Herbert and M.  Klein and E. Hodzic},
  journal={Manuscript submitted for publication},
  year={2025}
}

@article{chen_NeuralODE,
  title={Neural ordinary differential equations},
  author={Chen, Ricky TQ and Rubanova, Yulia and Bettencourt, Jesse and Duvenaud, David K},
  journal={Advances in neural information processing systems},
  volume={31},
  year={2018}
}

@book{GLM_marchenko2011sturm,
  title={Sturm-Liouville operators and applications},
  author={Marchenko, Vladimir Aleksandrovich},
  volume={373},
  year={2011},
  publisher={American Mathematical Soc.}
}

@article{GLM_gel1951determination,
  title={On the determination of a differential equation from its spectral function},
  author={Gel'fand, Izrail Moiseevich and Levitan, Boris Moiseevich},
  journal={Izvestiya Rossiiskoi Akademii Nauk. Seriya Matematicheskaya},
  volume={15},
  number={4},
  pages={309--360},
  year={1951},
  publisher={Russian Academy of Sciences, Steklov Mathematical Institute of Russian~…}
}

@inproceedings{KP_kadomtsev1970stability,
  title={On the stability of solitary waves in weakly dispersing media},
  author={Kadomtsev, Boris Borisovich},
  booktitle={Sov. Phys. Dokl.},
  volume={15},
  pages={539--541},
  year={1970}
}

@article{lax1968integrals,
  title={Integrals of nonlinear equations of evolution and solitary waves},
  author={Lax, Peter D},
  journal={Communications on pure and applied mathematics},
  volume={21},
  number={5},
  pages={467--490},
  year={1968},
  publisher={Wiley Online Library}
}

@book{IST_novikov1984theory,
  title={Theory of solitons: the inverse scattering method},
  author={Novikov, S and Manakov, Sergei V and Pitaevskii, Lev Petrovich and Zakharov, Vladimir Evgenevi{\v{c}}},
  year={1984},
  publisher={Springer Science \& Business Media}
}

@article{IST_gardner1967method,
  title={Method for solving the Korteweg-deVries equation},
  author={Gardner, Clifford S and Greene, John M and Kruskal, Martin D and Miura, Robert M},
  journal={Physical review letters},
  volume={19},
  number={19},
  pages={1095},
  year={1967},
  publisher={APS}
}

@book{IST_ablowitz1981solitons,
  title={Solitons and the inverse scattering transform},
  author={Ablowitz, Mark J and Segur, Harvey},
  year={1981},
  publisher={SIAM}
}

@article{kdv1895,
  title={XLI. On the change of form of long waves advancing in a rectangular canal, and on a new type of long stationary waves},
  author={Korteweg, Diederik Johannes and De Vries, Gustav},
  journal={The London, Edinburgh, and Dublin Philosophical Magazine and Journal of Science},
  volume={39},
  number={240},
  pages={422--443},
  year={1895},
  publisher={Taylor \& Francis}
}

@article{gomez2017reversible,
  title={The reversible residual network: Backpropagation without storing activations},
  author={Gomez, Aidan N and Ren, Mengye and Urtasun, Raquel and Grosse, Roger B},
  journal={Advances in neural information processing systems},
  volume={30},
  year={2017}
}

@article{Yu_Koop2025,
  title={Koopman theory-inspired method for learning time advancement operators in unstable flame front evolution},
  author={R. Yu and M. Herbert and M.  Klein and E. Hodzic},
	journal = {Physics of Fluids},
	volume = {37},
	pages = {024115},
	year = {2025}
}

@article{koopman1931hamiltonian,
  title={Hamiltonian systems and transformation in Hilbert space},
  author={Koopman, B.O.},
  journal={Proceedings of the National Academy of Sciences},
  volume={17},
  number={5},
  pages={315--318},
  year={1931},
  publisher={National Acad Sciences}
}

@book{brunton2022data,
  title={Data-driven science and engineering: Machine learning, dynamical systems, and control},
  author={Brunton, S.L. and Kutz, J.N.},
  year={2022},
  publisher={Cambridge University Press}
}

@article{mezic2013analysis,
  title={Analysis of fluid flows via spectral properties of the Koopman operator},
  author={Mezi{\'c}, I.},
  journal={Annual review of fluid mechanics},
  volume={45},
  number={1},
  pages={357--378},
  year={2013},
  publisher={Annual Reviews}
}

@article{kassam2005fourth,
  title={Fourth-order time-stepping for stiff PDEs},
  author={Kassam, Aly-Khan and Trefethen, Lloyd N},
  journal={SIAM Journal on Scientific Computing},
  volume={26},
  number={4},
  pages={1214--1233},
  year={2005},
  publisher={SIAM}
}

@article{rasool2021effect,
  title={Effect of non-ambient pressure conditions and Lewis number variation on direct numerical simulation of turbulent Bunsen flames at low turbulence intensity},
  author={Rasool, Raheel and Chakraborty, Nilanjan and Klein, Markus},
  journal={Combustion and Flame},
  volume={231},
  pages={111500},
  year={2021},
  publisher={Elsevier}
}

@article{YHN2024,
  title={Learning Flame Evolution Operator under Hybrid Darrieus Landau and Diffusive Thermal Instability},
  author={R. Yu and E. Hodzic and KJ Nogenmyr},
  journal={Energies},
  volume={17},
  number={13},
  pages={3097},
  year={2024},
  publisher={MDPI}
}

@article{YuH2024,
title = {Parametric learning of time-advancement operators for unstable flame evolution},
journal = {Physics of Fluids},
volume = {36},
pages = {044109},
year = {2024},
author = {R. Yu and E. Hodzic}
}

@article{Yu2023,
title = {Deep learning of nonlinear flame fronts development due to Darrieus–Landau instability},
journal = {APL Machine Learning},
volume = {1},
number = {2},
pages = {026106},
year = {2023},
author = {R. Yu}
}

@article{YBB15PRE,
author = {Yu, R. and Bai, X.S. and Bychkov, V.},
journal = {Phys. Rev. E},
number = {6},
pages = {063028},
title = {Fractal flame structure due to the hydrodynamic Darrieus-Landau instability},
volume = {92},
year = {2015}
}

@article{DARRIEUS1938UNPB,
author = {G. Darrieus},
journal={Unpublished work presented at La Technique Moderne},
title = {Propagation d’un front de flamme},
year = {1938},
}

@incollection{landau1988theory,
  title={On the theory of slow combustion},
  author={Landau, L},
  booktitle={Dynamics of curved fronts},
  pages={403--411},
  year={1988},
  publisher={Elsevier}
}

@incollection{zeldovich1944selected,
  title={Theory of Combustion and Detonation of Gases},
  author={YB Zeldovich},
  booktitle={Selected Works of Yakov Borisovich Zeldovich, Volume I: Chemical Physics and Hydrodynamics},
  year={1944},
  publisher={Princeton University Press}
}

@article{sivashinsky1977diffusional,
  title={Diffusional-thermal theory of cellular flames},
  author={Sivashinsky, GI},
  journal={Combustion Science and Technology},
  volume={15},
  number={3-4},
  pages={137--145},
  year={1977},
  publisher={Taylor \& Francis}
}

@article{CRETA2011INST,
author = {F. Creta and N. Fogla and M. Matalon},
title = {Turbulent propagation of premixed flames in the presence of Darrieus–Landau instability},
journal = {Combustion Theory and Modelling},
volume = {15},
number = {2},
pages = {267-298},
year  = {2011},
}

@article{Creta2020propagation,
  title={Propagation of premixed flames in the presence of Darrieus--Landau and thermal diffusive instabilities},
  author={Creta, Francesco and Lapenna, Pasquale Eduardo and Lamioni, Rachele and Fogla, Navin and Matalon, Moshe},
  journal={Combustion and Flame},
  volume={216},
  pages={256--270},
  year={2020},
  publisher={Elsevier}
}

@article{Karlin2002cellular,
  title={Cellular flames may exhibit a non-modal transient instability},
  author={Karlin, V},
  journal={Proceedings of the Combustion Institute},
  volume={29},
  number={2},
  pages={1537--1542},
  year={2002},
  publisher={Elsevier}
}

@article{denet2006stationary,
  title={Stationary solutions and Neumann boundary conditions in the Sivashinsky equation},
  author={Denet, Bruno},
  journal={Physical Review E},
  volume={74},
  number={3},
  pages={036303},
  year={2006},
  publisher={APS}
}

@article{Thual_Frisch_Henon_poledecomp,
author = {O. Thual and U. Frisch and M. Hénon},
title = {Application of pole decomposition to an equation governing the dynamics of wrinkled flame fronts },
journal = {Journal de Physique},
volume = {46},
pages = {1485-1494},
year = {1985},
}

@book{Kupervasser_pole_book,
author = {Kupervasser, O.},
publisher = {Springer International Publishing},
title = {{Pole Solutions for Flame Front Propagation}},
year = {2015}
}

@article{Vaynblat_matalon_polestability1,
author = { D. Vaynblat and M. Matalon},
title = {  Stability of Pole Solutions for Planar Propagating Flames: I. Exact Eigenvalues and Eigenfunctions },
journal = { SIAM J.  Appl. Math.},
volume = {60},
pages = {679-702},
year = {2000},
}

@article{Vaynblat_matalon_polestability2,
author = { D. Vaynblat and M. Matalon},
title = { Stability of Pole Solutions for Planar Propagating Flames: II. Properties of Eigenvalues/Eigenfunctions and Implications to Stability},
journal = { SIAM J.  Appl. Math.},
volume = {60},
pages = {703-728},
year = {2000},
}

@article{Olami_noise,
author = {Z Olami and B Galanti and O Kupervasser and I Procaccia},
title = {Random noise and pole dynamics in unstable front propagation},
journal = {Physical Review E },
volume = {55},
pages = {2649},
year = {1997}
}

@article{kuramoto1978diffusion,
  title={Diffusion-induced chaos in reaction systems},
  author={Kuramoto, Yoshiki},
  journal={Progress of Theoretical Physics Supplement},
  volume={64},
  pages={346--367},
  year={1978},
  publisher={Oxford University Press}
}

@article{SivaEq,
author = {G.I.Sivashinsky},
title = {Nonlinear analysis of hydrodynamic instability in laminar flames—I. Derivation of basic equations},
journal = {Acta Astronautica},
volume = {4},
pages = {1177-1206},
year = {1977},
}

@article{michelson1977nonlinear,
  title={Nonlinear analysis of hydrodynamic instability in laminar flames—II. Numerical experiments},
  author={Michelson, Daniel M and Sivashinsky, Gregory I},
  journal={Acta astronautica},
  volume={4},
  number={11-12},
  pages={1207--1221},
  year={1977},
  publisher={Elsevier}
}

@InProceedings{CNN1,
author = {X. Guo and W. Li and  F. Iorio},
title = { Convolutional neural networks for steady ﬂow approximation},
booktitle={Proceedings of the 22nd ACM SIGKDD International Conference on Knowledge Discovery and Data Mining},
year = {2016},
}

@article{CNN2,
author = {Zhu, Y. and Zabaras, N.},
title = {Bayesian deep convolutional encoder–decoder networks for surrogate modeling and uncertainty quantification},
journal = {Journal of Computational Physics},
volume = {366},
pages = {415-447},
year = {2018},
}

@article{CNN3,
author = {Adler, J. and Öktem, O.},
title = {Solving ill-posed inverse problems using iterative deep neural networks},
journal = {Inverse Problems},
volume = {33},
pages = {124007},
year = {2017},
}

@article{CNN4,
author = {Bhatnagar, S. and Afshar, Y. and Pan, S. and Duraisamy, K. and Kaushik, S.},
title = {Prediction of aerodynamic flow fields using convolutional neural networks},
journal = {Computational Mechanics},
volume = {64},
pages = {525-545},
year = {2019},
}

@article{CNN5,
 title={Solving parametric PDE problems with artificial neural networks},
 volume={32},
 DOI={10.1017/S0956792520000182}, 
 number={3}, 
 journal={European Journal of Applied Mathematics}, 
 publisher={Cambridge University Press}, 
 author={Khoo, YUEHAW and Lu, JIANFENG and Ying, LEXING}, 
 year={2021}, 
 pages={421–435}
 }

@Unpublished{UNet,
author = {O. Ronneberger and P. Fischer and T. Brox},
title = {U-Net: Convolutional Networks for Biomedical Image Segmentation},
year = {2015},
archivePrefix = {arXiv},
eprint = {1505.04597},
primaryClass = {cs.CV},
}

@article{ConvPDE,
title = {ConvPDE-UQ: Convolutional neural networks with quantified uncertainty for heterogeneous elliptic partial differential equations on varied domains},
journal = {Journal of Computational Physics},
volume = {394},
pages = {263-279},
year = {2019},
issn = {0021-9991},
doi = {https://doi.org/10.1016/j.jcp.2019.05.026},
url = {https://www.sciencedirect.com/science/article/pii/S0021999119303572},
author = {Nick Winovich and Karthik Ramani and Guang Lin},
}

@Unpublished{GraphKenerlNetwork,
author = {Z. Li and N. Kovachki and K. Azizzadenesheli and B. Liu and K. Bhattacharya and A. Stuart and A. Anandkumar},
title = {Neural Operator: Graph Kernel Network for Partial Differential Equations},
year = {2020},
archivePrefix = {arXiv},
eprint = {2003.03485},
primaryClass = {cs.LG},
}

@article{kovachki2021neural,
  title={Neural operator: Learning maps between function spaces},
  author={Kovachki, Nikola and Li, Zongyi and Liu, Burigede and Azizzadenesheli, Kamyar and Bhattacharya, Kaushik and Stuart, Andrew and Anandkumar, Anima},
  journal={arXiv preprint arXiv:2108.08481},
  year={2021}
}

@Unpublished{FNO2020,
author = {Li, Z. and Kovachki, N.  and Azizzadenesheli, K.  and Liu, B.  and Bhattacharya, K.  and Stuart, A. and Anandkumar, A.},
title = {Fourier neural operator for parametric partial differential equations},
year = {2020},
archivePrefix = {arXiv},
eprint = {2010.08895},
primaryClass = {cs.LG},
}

@article{gupta2021multiwavelet,
  title={Multiwavelet-based operator learning for differential equations},
  author={Gupta, Gaurav and Xiao, Xiongye and Bogdan, Paul},
  journal={Advances in neural information processing systems},
  volume={34},
  pages={24048--24062},
  year={2021}
}

@article{tripura2023wavelet,
  title={Wavelet neural operator for solving parametric partial differential equations in computational mechanics problems},
  author={Tripura, Tapas and Chakraborty, Souvik},
  journal={Computer Methods in Applied Mechanics and Engineering},
  volume={404},
  pages={115783},
  year={2023},
  publisher={Elsevier}
}

@Unpublished{DeepONet,
author = {L. Lu and P. Jin and G.E. Karniadakis},
title = {DeepONet: Learning nonlinear operators for identifying differential equations based on the universal approximation theorem of operators},
year = {2019},
archivePrefix = {arXiv},
eprint = {1910.03193},
primaryClass = {cs.LG},
}

@article{chen2023laplace,
  title={Laplace neural operator for complex geometries},
  author={Chen, Gengxiang and Liu, Xu and Li, Yingguang and Meng, Qinglu and Chen, Lu},
  journal={arXiv preprint arXiv:2302.08166},
  year={2023}
}

\end{document}